\documentclass[letterpaper, 10 pt, conference]{IEEEconf}  

\usepackage{graphicx}      
\usepackage{subcaption} 
\usepackage{amsmath,amssymb}
\usepackage{cite}          
\usepackage{hyperref}      
\usepackage{xcolor}
\usepackage{listings}
\usepackage{comment}

\newcommand{\invisible}[1]{#1} 

\newcommand{\nb}[1]{} 

\newcommand{\anonym}[1]{#1} 

\newcommand{\anocite}[1]{#1} 

\title{\LARGE\bf
Pogobot: an Open-Source, Low-Cost Robot \\ for Swarm Robotics and Programmable Active Matter
}

\invisible{
\author{
Alessia Loi\textsuperscript{1},
Loona Macabre\textsuperscript{1},
Jérémy Fersula\textsuperscript{1,2}, 
Keivan Amini\textsuperscript{1,2},
Leo Cazenille\textsuperscript{1},
Fabien Caura\textsuperscript{1},
\\
Alexandre Guerre\textsuperscript{3},
Stéphane Gourichon\textsuperscript{1},
Laurent Fabre\textsuperscript{1},
Olivier Dauchot\textsuperscript{2},
Nicolas Bredeche\textsuperscript{1,4*}
\\
\bigskip
\\
(1) Sorbonne Universit\'{e}, CNRS, ISIR, F-75005 Paris, France
\\
(2) Gulliver Lab, ESPCI Paris - PSL, Paris, France
\\
(3) Sorbonne Universit\'{e}, SUMMIT, F-75005 Paris, France
\\
(4) Sorbonne Universit\'{e}, CNRS, IBPS, LJP, F-75005 Paris, France
\\
\bigskip
\\
* correspondence: nicolas.bredeche@sorbonne-universite.fr
}
}

\begin{document}

\maketitle

\pagestyle{plain} 


\begin{abstract}

This paper describes the Pogobot, an open-source platform specifically designed for research at the interface of swarm robotics and active matter. Pogobot features vibration-based or wheel-based locomotion, fast infrared communication, and an array of sensors in a cost-effective package (approx. 250~euros/unit). The platform's modular design, comprehensive API, and extensible architecture facilitate the implementation of swarm intelligence algorithms and collective motion. Pogobots offer an accessible alternative to existing platforms while providing advanced capabilities including directional communication between units and fast locomotion, all with a compact form factor. More than $200$ Pogobots are already being used on a daily basis in several Universities to study self-organizing systems, programmable active matter, discrete reaction-diffusion-advection systems and computational models of social learning and evolution. This paper details the hardware and software architecture, communication protocols, locomotion mechanisms, and the infrastructure built around the Pogobots\nb{, and demonstrates complex collective behaviors implemented on a large-scale swarm of Pogobots, validating the platform's effectiveness}.

\end{abstract}






\begin{figure}[!b]
    \centering
    \includegraphics[height=3cm]{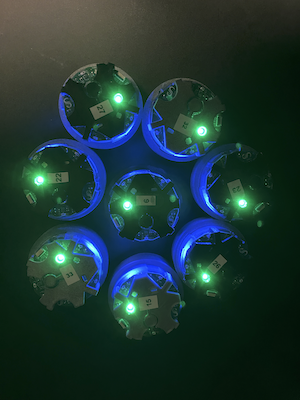}
    \hspace{0.1cm}%
    \includegraphics[height=3cm]{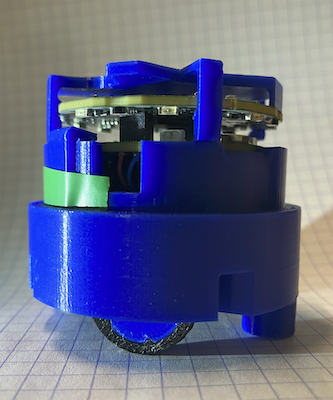}%
    \hspace{0.1cm}%
    \includegraphics[height=3cm]{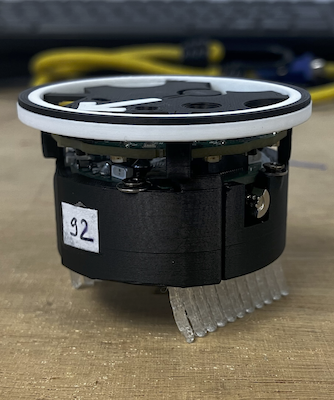}%
    \hspace{0.1cm}%
    \includegraphics[height=3.5cm]{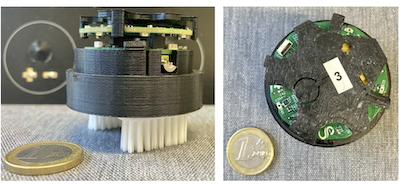}%
    \caption{From top-left to bottom-right: (a) a small swarm of $8$ Pogobots; (b) Pogobot with wheel-based locomotion;  Pogobot with vibration-induced locomotion, using (c) TPU 3D-printed brushes or (d) or commercial toothbrushes head with or without inclined brush ; (e) view from above (the Pogobot is $\sim$6~cm diameter).}
    \label{fig:pogobots}
\end{figure}

\section{Introduction}

We present the Pogobot robot, a robotic platform developed to provide an accessible, open-source, open-hardware and cost-effective robotic solution, specifically designed to support experimentation in swarm and collective robotics. Its primary goal is to offer researchers a scalable and adaptable hardware platform that enables large-scale studies without significant financial barriers. 

The Pogobot started as a joint project between \anonym{Sorbonne Université} and \anonym{ESPCI Paris-PSL} in \anonym{France}, and is now also used in two other institutes (\anonym{Université de Lorraine} and \anonym{Ochanomizu University} in \anonym{Japan}). While Pogobot-based research is still young, part or all of the elements of the Pogobot design have already been used in published research. Firstly, preliminary designs of the locomotion modes were used in works studying the interaction between social learning and steric interactions~\anocite{\cite{zion2023morphological}}, as well as for rigorously characterizing the dynamics of using commercial toothbrush vibration-based locomotion~\anocite{\cite{fersula2024self}}. Secondly, a swarm of fully functional Pogobots have been used to study the dynamics of diffusion in self-organized systems for collective decision making~\anocite{\cite{cazenille2024hearing}}, pattern formation~\anocite{\cite{loi2025evolving}} and consensus in highly motile swarms.

The Pogobot, as illustrated in Fig.~\ref{fig:pogobots}, is intended as a basic component of a model swarm system, thanks to its accessibility, and versatility in terms of programming (compiled C code running a softcore RISC-V CPU running on an FPGA), changeable locomotion scheme (wheeled or vibration-based stick-slip), 3D-printable form factor (round shape or other), and multi-directional fast IR-based communication system. Pogobots can be programmed one by one using a physical cable or as a collective using an overhead IR controller, inspired by the Kilobot platform~\cite{rubenstein2014programmable}, to which Pogobots are intended to be a modern replacement. 

In this paper, we provide a technical description of the Pogobot and its ecosystem, starting with hardware (Section~\ref{sec:design}) and software (Section~\ref{sec:sw}) specifications. We then present and evaluate the efficiency of the locomotion schemes (Section~\ref{sec:loco}) ad the communication architecture (Section~\ref{sec:com}). Finally, we present the various software support tools and hardware extensions that are part of the wider Pogobot ecosystem (Section~\ref{sec:ext}).

\begin{table*}[t]
\centering
\small
\renewcommand{\arraystretch}{1.1}
\setlength{\tabcolsep}{2pt}
\begin{tabular}{>{\raggedright\arraybackslash}p{0.22\textwidth}!{\vrule width 0.2pt}
                >{\raggedright\arraybackslash}p{0.08\textwidth}!{\vrule width 0.2pt}
                >{\raggedright\arraybackslash}p{0.17\textwidth}!{\vrule width 0.2pt}
                >{\raggedright\arraybackslash}p{0.08\textwidth}!{\vrule width 0.2pt}
                >{\raggedright\arraybackslash}p{0.10\textwidth}!{\vrule width 0.2pt}
                >{\raggedright\arraybackslash}p{0.11\textwidth}!{\vrule width 0.2pt}
                >{\raggedright\arraybackslash}p{0.04\textwidth}!{\vrule width 0.2pt}
                >{\raggedright\arraybackslash}p{0.10\textwidth}!{\vrule width 0.2pt}
                >{\raggedright\arraybackslash}p{0.09\textwidth}
                }

\hline
\textbf{Platform} & \textbf{Research Field} & \textbf{Footprint} & \textbf{Locomot.} & \textbf{Sensors} & \textbf{Pairwise Comm.} & \textbf{CPU} & \textbf{Assembled cost} & \textbf{Open} \\ 
\hline

\textbf{Pogobot (ours)} & SR+PAM & Ø 55~mm, $\sim$43-56g & Wheels or vibration & IR, 3x light, IMU & fast IR (1–10~MHz) & yes & $\sim$250€ & yes \\ 

e-puck2~\cite{mondada2009puck} & SR & Ø 70~mm, $\sim$150g & Wheels & Camera, IR & IR, Bluetooth & yes & $\sim$1000€$^*$ & yes \\ 

Khepera~\cite{mondada1999development,prorok2010indoor} & SR & Ø 55-140mm, $\sim$540g & Wheels & IR & IR + add-ons & yes & $\sim$3000€$^*$ & no \\ 

Elisa-3~\cite{Elisa3} & SR & Ø 50~mm, $\sim$40g & Wheels & IR, light & IR + RF & yes & $\sim$400€$^*$ & yes \\ 



S-bot~\cite{Mondada2005Swarmbot} & SR & Ø 190~mm, $\sim$700g & Wheels & Camera, IR & IR, vision & yes & unknown & no \\ 


Kilobot~\cite{Rubenstein2014Kilobot} & SR & Ø 33~mm, $\sim$16g & Vibration & 1x IR, light & 1x IR & yes & $\sim$100€$^{*\dagger}$ & yes \\

Polar disks~\cite{Deseigne2010PRL,Deseigne2012PRE,Anderson2024Fronts} & AM & Ø $\sim$4~mm, $\sim$0.1g & Vibration & \textit{none} & \textit{none} & no & $\sim$1€ & yes \\ 

Bristlebots/Hexbug~\cite{Deblais2018PRL,VolpeAJP2024,Baconnier2022NatPhys,boudet2022effective,Xi2024PNAS} & AM & Ø $\sim$45~mm, $\sim$7g & Vibration & \textit{none} & \textit{none} & no & $\sim$5€ & yes (DIY)\\ 

Swarmodroid~\cite{dmitriev2023swarmodroid} & PAM & Ø $\sim$50~mm, $\sim$20~g & Vibration & \textit{none} & \textit{none} & yes & unknown & yes \\

GRASPion~\cite{novkoski2025graspion} & PAM & $\sim$60x30~mm, $\sim$17~g & Vibration & IR, Mag & \textit{none} & yes & unknown & yes \\

\end{tabular}
\caption{
Swarm Robotics (SR), Active Matter (AM) and Programmable Active Matter (PAM) platforms. The assembled cost is considered instead of the raw cost (e.g. Kilobots raw component cost is announced at $\sim$14€, Swarmodroid at $\sim$11€ and GRASPion at $\sim$15€) to account for the true cost paid when acquiring the robot (i.e., building it or buying it from a vendor). A star $^*$ indicates the market price, if any. A dagger $^\dagger$ is for when marketing option was discontinued.}
\label{tab:platforms}
\end{table*}

\section{Design and Hardware}
\label{sec:design}

\begin{figure*}[htbp]
    \centering
    \includegraphics[width=1.0\linewidth]{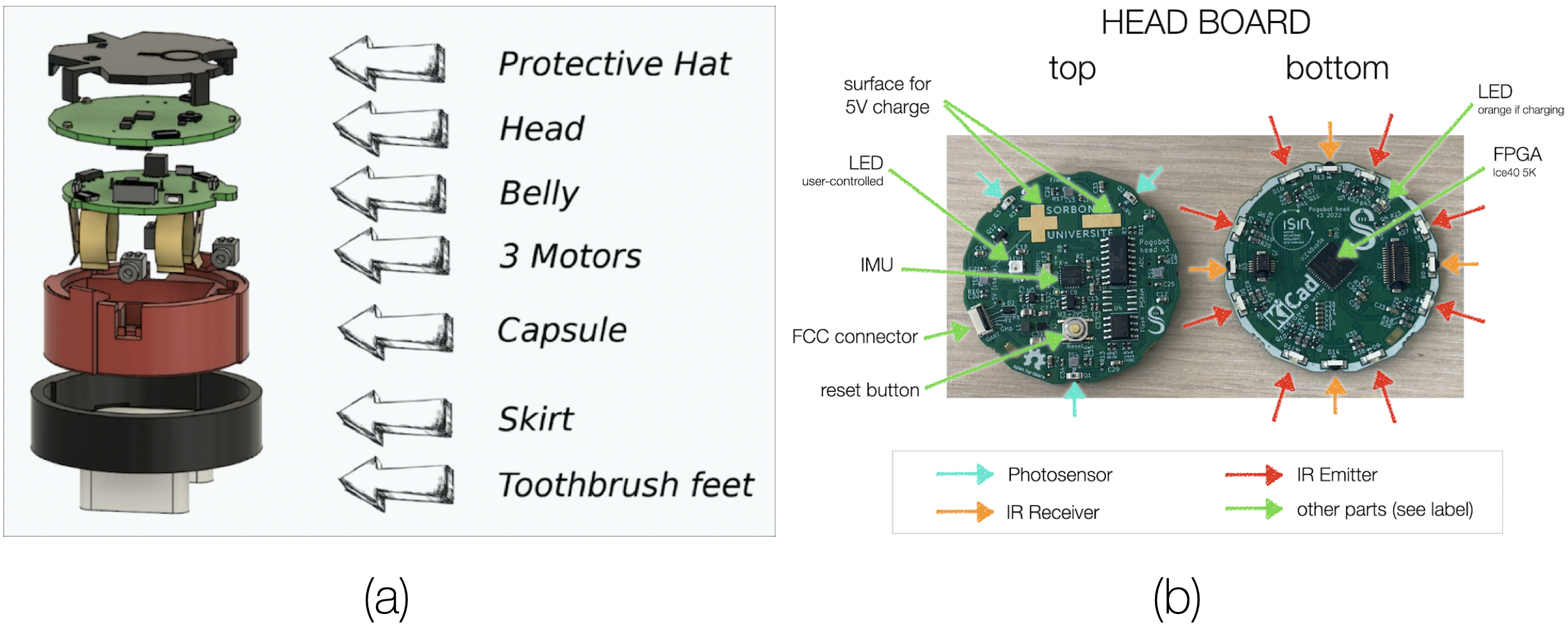}%
    \caption{(a) Schematic representation of the Pogobot showing its main components. (b) Close-up of the Head board.}
    \label{fig:pogobot_general}
\end{figure*}

As shown in 
Figure~\ref{fig:pogobot_general}-(a), 
Pogobot consists primarily of two printed circuit boards (PCBs), named \textbf{Head} and \textbf{Belly}.\invisible{(more details in Annex~\ref{annex:hw}.}
The Head and Belly boards are easily connected, and can be encased in a 3D-printed \textbf{Capsule}, which itself can be plugged into a \textbf{Skirt} that determines the external shape of the robot and embeds the actual locomotion device (depending on the model: 3D-printed brushes and/or legs, toothbrush heads or wheels). An easy-to-fit 3D-printed \textbf{Hat} on top the robot offers protection during handling. Aside from the Head, which requires specific skills available at any electronic manufacturing services, all other components can be designed in the FabLab. All that is required to build and run a Pogobot is available at~\anonym{\url{https://pogobot.github.io/}}, with \textit{everything} being open-source and open-hardware.

The \textbf{Head} board (see 
Fi~\ref{fig:pogobot_general}-(b)
) integrates essential computational and communication hardware, including a Lattice FPGA ICE40up5k~\cite{ice40} running a softcore RISC-V 32-bit microprocessor (VexRISC-V~\cite{vexriscv}, integrated into a SoC using LiteX~\cite{kermarrec2020litexopensourcesocbuilder}). It incorporates four infrared (IR) emitter/receiver pairs for omnidirectional communication (see Section~\ref{sec:com}). We use TS4231 chips~\cite{triad2016ts4231} to ensure fast IR communication (support 1-10MHz optical carrier frequencies, to be compared to approx. 38kHz for classic IR communication). Additionally, the Head board is equipped with three ambient light photosensors (front-right, front-left, back) for detecting light intensity and gradient, an inertial measurement unit (IMU), and two LEDs: one for user feedback and the other for indicating charging status. The HEAD features an FCC connector for programming and charging, and two contact-based surface for cable-free charging (The Pogobot can be simply placed on a charging pad, eliminating the need for cables or connectors).

The Belly board focuses on energy management and locomotion. It features three motor interfaces used to power three motors (two are used for differential drive, one is available for any further need). The exact use of the motor for motility varies depending on the design used (either vibration-enabled stick and slip, or wheeled locomotion, see Fig.\ref{fig:pogobots}). The Belly board also supports battery management and includes the structure for connecting a 3.3V LiFePO4 CR2 battery underneath. Using LiFePO4 technology was motivated by durability and safety features, as well as their lower environmental impact (incl. being cobalt-free). In addition, four programmable LEDs are positioned to face each direction. Autonomy is from 60~min (motors at full speed, all LEDs on) to 120~min (motors half-speed, no LED, no communication).

All 3D-printed parts (Capsule, Skirt, Hat) are customisable. The Capsule is designed as a universal cradle for the electronics, motors and battery. The Skirt is plugged underneath the Capsule to accommodate the physical part of the vibration-based locomotion: toothbrush heads, 3D-printed legs, or brushes can be attached (see Section~\ref{sec:loco}). The Capsule and Skirt can also be printed as one module, including a specific design for wheel-based locomotion (wheels are also 3D-printed).



Table~\ref{tab:platforms} shows how Pogobot takes a unique position at the intersection of Swarm Robotics and (Programmable) Active Matter. To some extent, Pogobot is envisioned as an improved replacement for the famous but discontinued Kilobot platform. Moreover, the versatile nature of the Pogobot 3D-printed parts, which can be re-designed to fit a particular set of requirements, is meant to be particularly attractive for experimental studies in soft matter and active matter physics.

\section{Software Architecture and API}
\label{sec:sw}

\begin{figure}[t]
\centering
\includegraphics[width=0.49\textwidth]{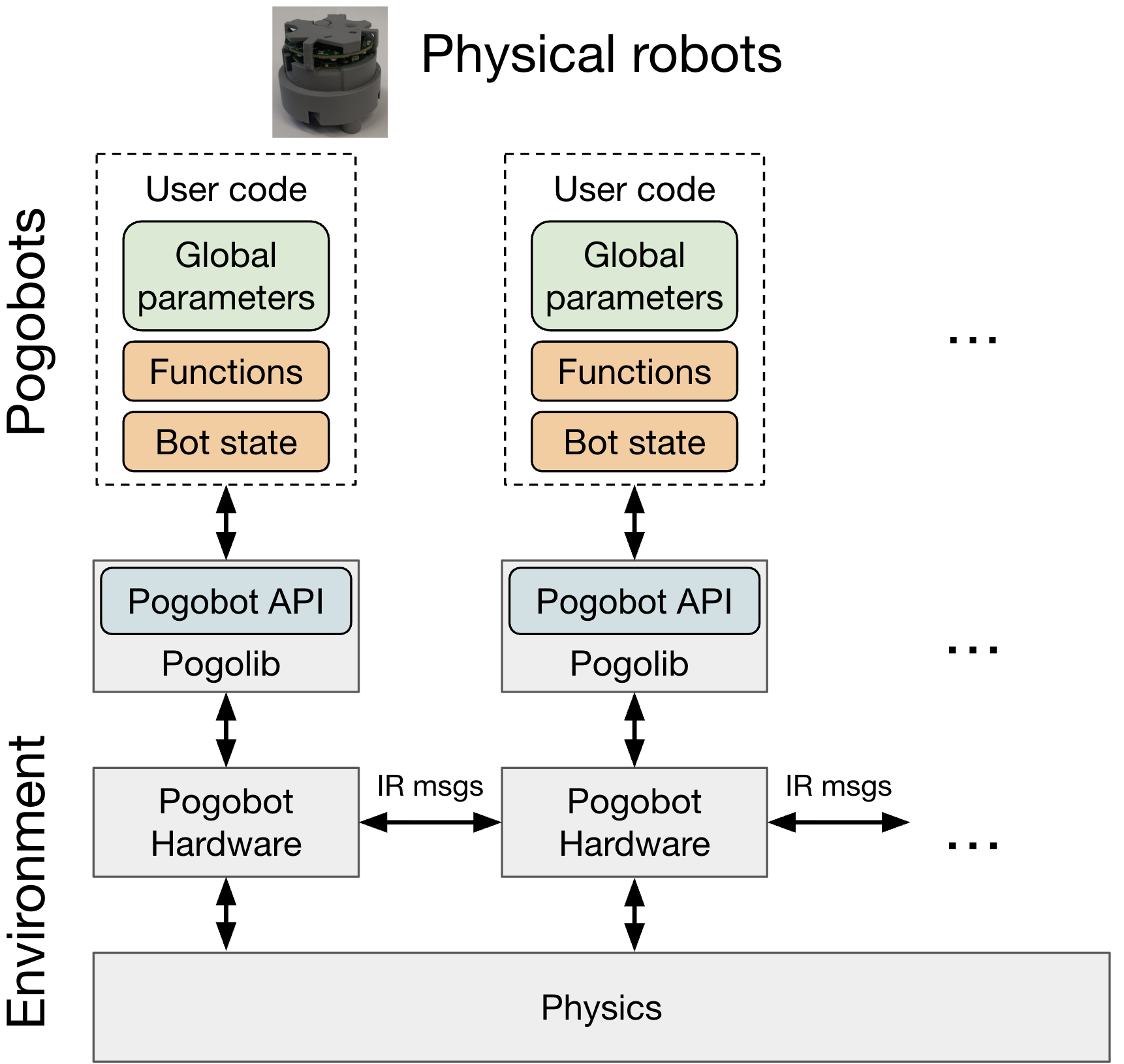}%
\caption{Pogobot software and hardware stack. Each robot combines three levels: user program, API (PogoLib/PogoAPI), and hardware. Robot-to-robot interactions is performed through IR communication.}
\label{fig:swarchi}
\end{figure}

The programming environment for Pogobot is designed to be user-friendly. Figure~\ref{fig:swarchi} illustrates the complete Pogobot's stack, and interaction with the world and other Pogobots. The Pogobot software stack is available on Github\footnote{ \anonym{\url{https://github.com/nekonaute/pogobot/blob/main/pogodocs.md}}}, and consists of (1) \textbf{Pogobios}, a low-level software interface which allows basic interactions such as configuration and programming, (2) \textbf{PogoLib}/\textbf{PogoAPI}: a high-level API for programming, and (3) User programs, written in C and compiled to run on the RISC-V softcore, including multiple demos and programming templates to start working with Pogobots\invisible{ (More details are in Annex~\ref{annex:sw}}.

It is possible to program Pogobots via direct cable connections or wirelessly using infrared-based technologies, such as with the PogoShower (see Section~\ref{sec:hwext}). The PogoShower device, inspired by Kilobot's overhead controller, enables simultaneous programming of multiple robots, significantly simplifying deployment in large-scale experimental settings.

\section{Locomotion}
\label{sec:loco}

\begin{figure*}[t]
  \centering
   \includegraphics[width=0.25\linewidth]{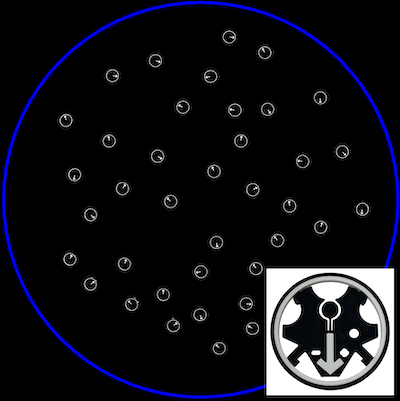} 
   \includegraphics[width=0.36\linewidth]{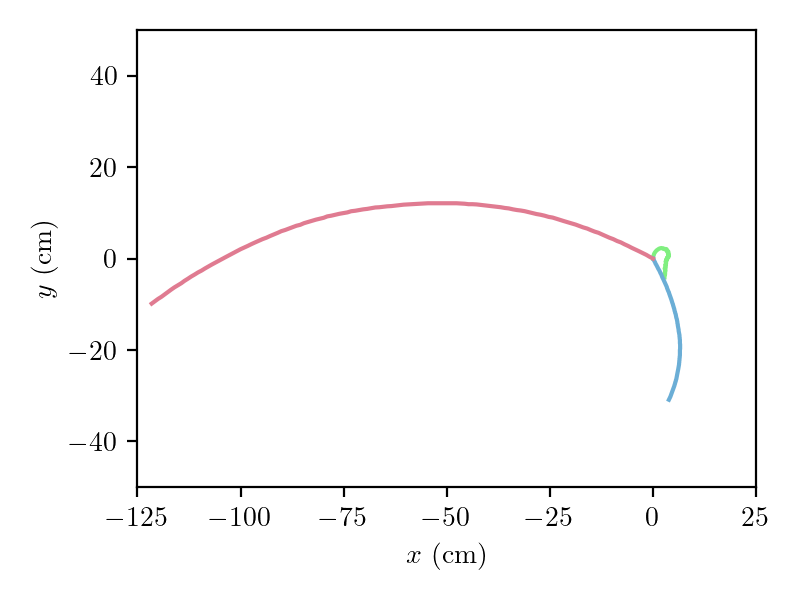} 
   \includegraphics[width=0.36\linewidth]{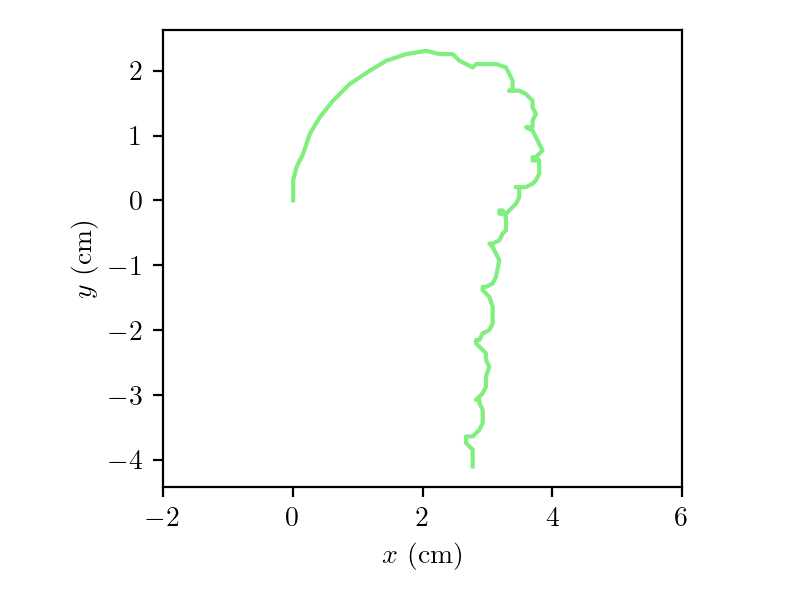} 
  \vspace{-3mm}  
  \caption{(a) Snapshot of an experiment in real conditions, with extracted position and orientation of all robots. Inset: a top view of the Hat used for tracking. (b) Example of trajectories for each locomotion modality (red: wheels; blue: 3d-printed TPU legs; green: commercial toothbrush legs) (c) zoom on the trajectory obtained with the toothbrush.}
  \label{fig:tracking}
\end{figure*}

\begin{figure*}[t]
  \centering
  \includegraphics[width=0.32\textwidth]{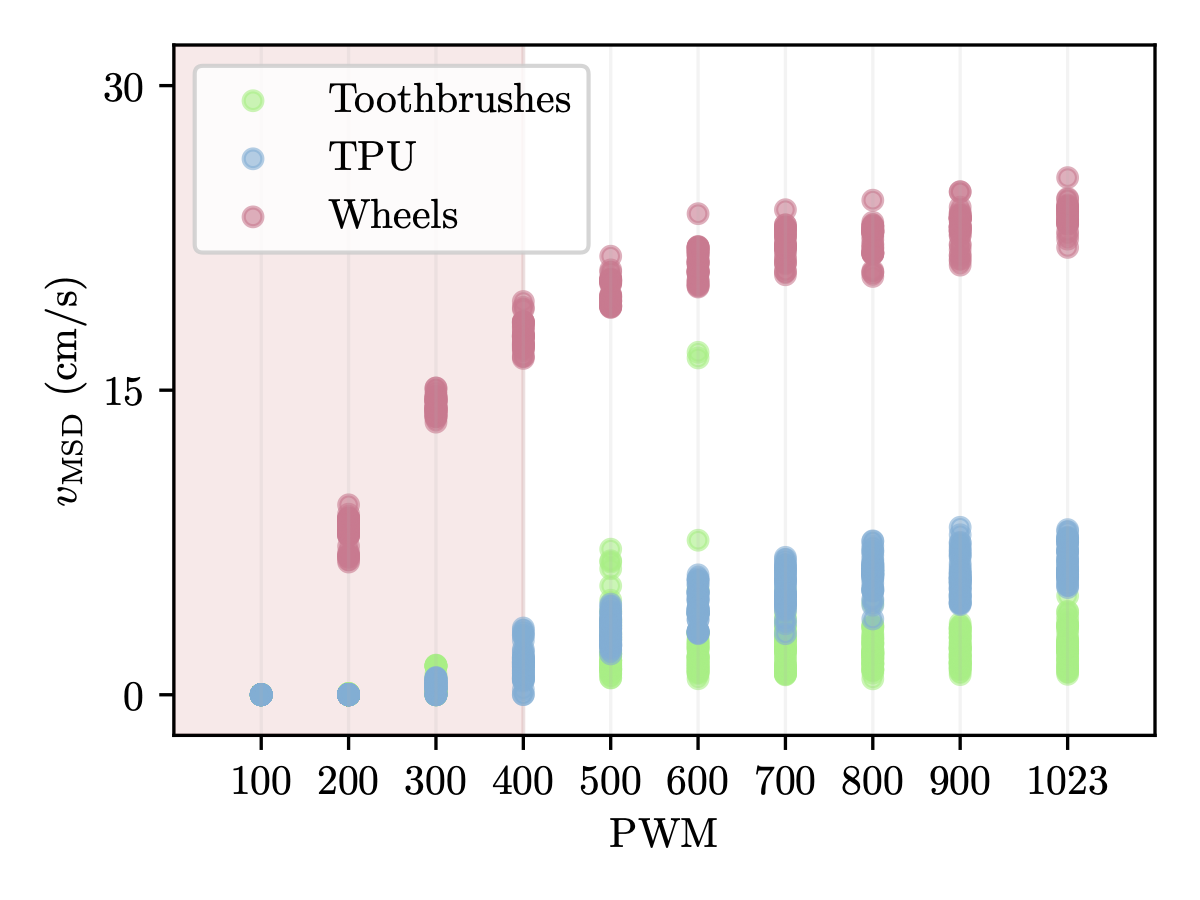}\hfill
  \includegraphics[width=0.32\textwidth]{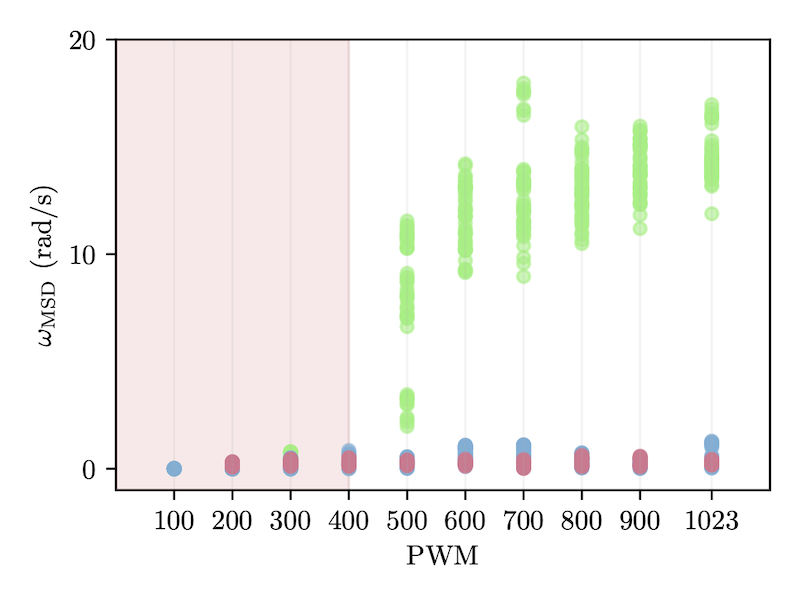}\hfill
  \includegraphics[width=0.32\textwidth,page=1]{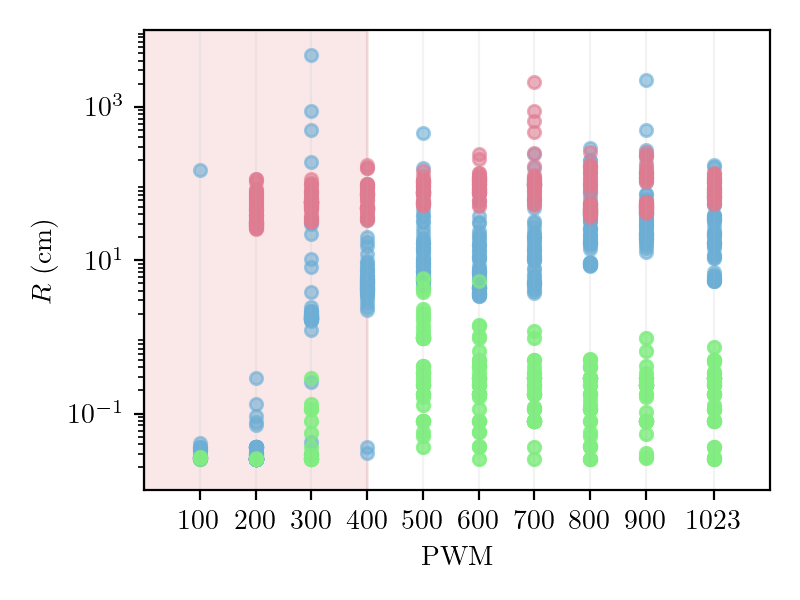}
  \vspace{-3mm}  
  \caption{Characterization of the dynamics observed for the three locomotion modalities reported above: (a) speed of the center, (b) angular frequency of the orientation and (c) radius of curvature obtained for increasing values of PWM.}
  \label{fig:tpu_tooth_wheels}
\end{figure*}

 A major advantage of the Pogobot platform is its high degree of customizability. In this Section, we provide a dynamical characterization of two locomotion modalities: wheel-based Pogobots (Fig.~\ref{fig:pogobots}-a) and vibration-induced Pogobots (Fig.~\ref{fig:pogobots}-b\&c). In the latter case, two different toothbrushes materials are considered: a classical one in nylon (commercial toothbrush head) without any inclination, and a 3D-printed inclined brush made of TPU 82A. TPU brushes are inclined at $10^\circ$ in order to increase the forward driving force and to improve straight-line motion.


Experiments are carried out in a circular arena of radius $r = 75$ cm, placed on a flat matte-black surface. The arena is surrounded by black cardboard walls and lighten using led strips to ensure homogeneous illumination. A 4096 x 3000 pixels RGB camera is mounted above the arena and capture the Pogobot's motions. A custom 3D-printed Hat with a black-and-white annulus (see  Figure~\ref{fig:tracking}-a) enables reliable measurements of the center $(x,y)$ and orientation $\theta$ of the Pogobots.

For both vibration-driven and wheel-equipped Pogobots, 5 robots are programmed via the Pogobot C API to run at prescribed rotation frequency of their motors as imposed by the Pulse Width Modulation (PWM) values. The two motors are given equal PWM values. The wheel-equipped Pogobots then follow an automatic calibration procedure using the onboard inertial measurement unit and implementing a Kalman filter in order to minimize systematic biases and improve straight-line motion (See Annex~\anonym{\ref{sec:autocalibration}}).

For each PWM value, each Pogobot is run for $6$ s over $10$ independent trials. Each experiment is recorded and subsequently processed and analyzed using the Pogotrack software (see Section~\ref{sec:pogotracker}).  
A dedicated Pogotrack module removes trajectory segments affected by wall collisions and extracts three physical observables: the linear velocity $v$ of the center, the angular velocity of the orientation $\Omega$, and the curvature radius of the trajectory $R$.

Experimental observations highlight clear differences between locomotion designs, as seen from the trajectories (see Fig.~\ref{fig:tracking}-b). More quantitatively (see Fig.~\ref{fig:tpu_tooth_wheels}), the wheels equipped Pogobots perform the fastest and straightest motion, with speed reaching up to $25$ cm/s and the trajectories presenting a radius of curvature $R\sim 100$ cm. By comparison, the TPU legs Pogobots, which for the purpose of illustration are here not calibrated, only reach speeds up to $7$ cm and exhibit stronger bias, with radius of curvature $R\sim 10$ cm. Finally, the Nylon brushes primarily induce spinning, with angular frequency reaching $15$~rad/s. We also note that wheel-based robots display the most consistent dynamics across individuals, whereas nylon toothbrush legs show the largest variability.

\section{Communication}
\label{sec:com}

\begin{figure*}[t]
\centering
\includegraphics[width=0.49\textwidth]{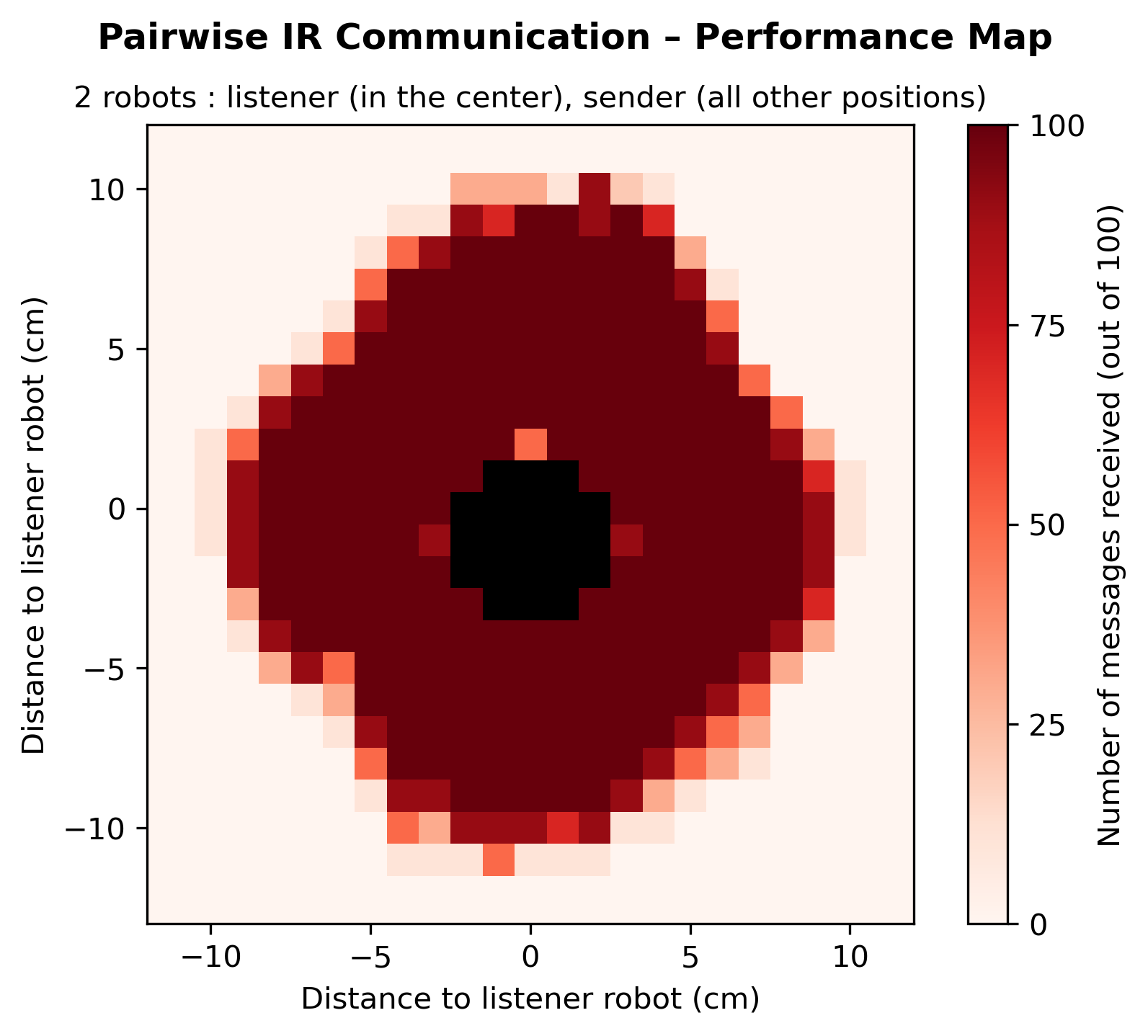}%
\hfill
\includegraphics[width=0.474\textwidth]{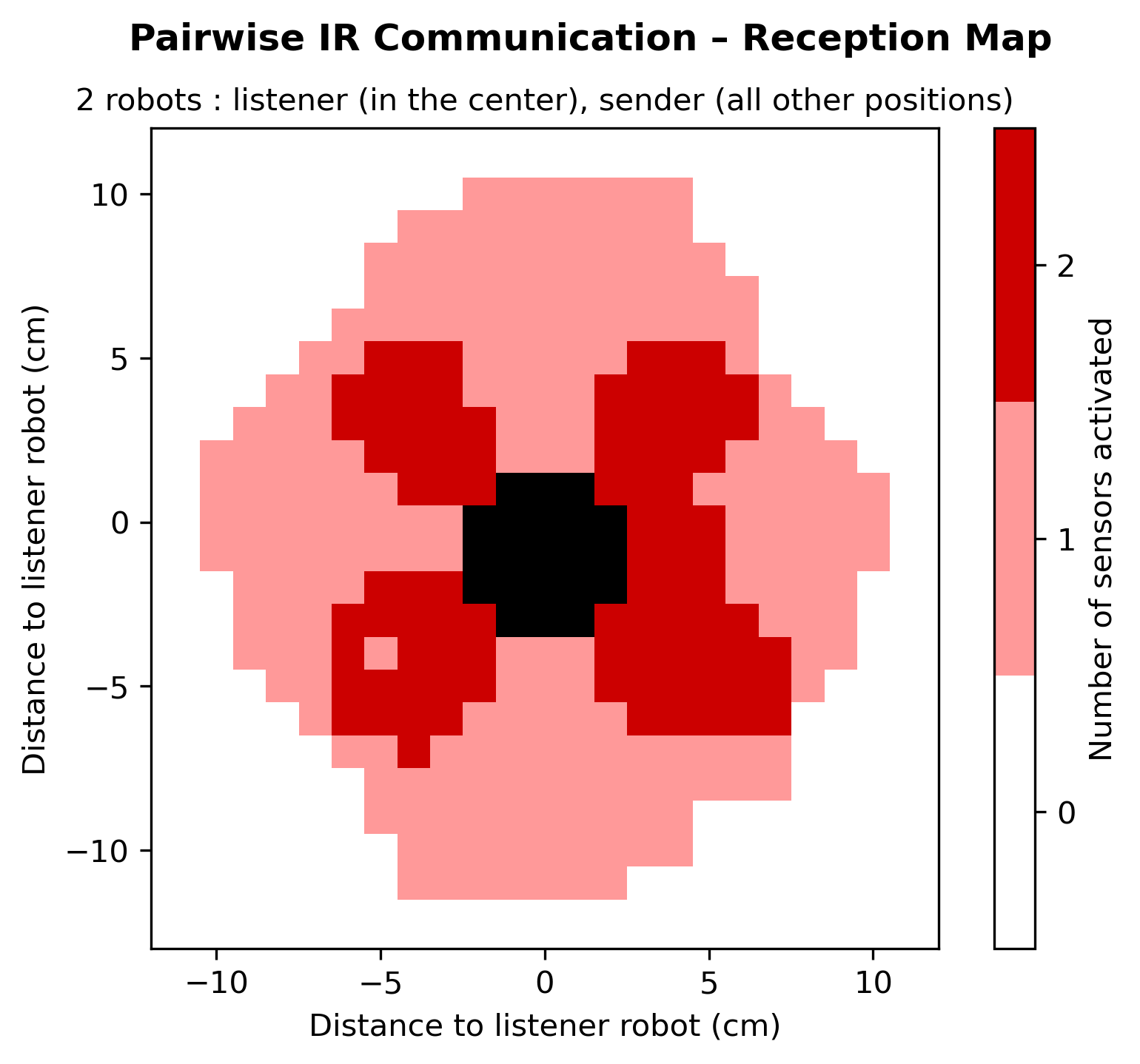}
\caption{Performance (a) and reception (b) maps of pairwise IR communication. The focal robot (black cells in the center) listens; a sender robot transmits with its front emitter. The sender is manually positioned to test communication from all positions around the listener (resolution of 1 cm$^2$ (left) or 0.5cm$^2$ (right)), always facing the receiver, until it is out of range. Colors indicate the number of messages successfully received out of 100 trials (0 = white, 100 = dark red). Black cells correspond to the focal robot’s footprint, excluded from measurement. Data shown for two typical robots, compiled from 100 messages (payload of 64 bytes).}
\label{fig:IRmaps}
\end{figure*}

\begin{figure*}
    \centering
    \includegraphics[width=0.33\linewidth]{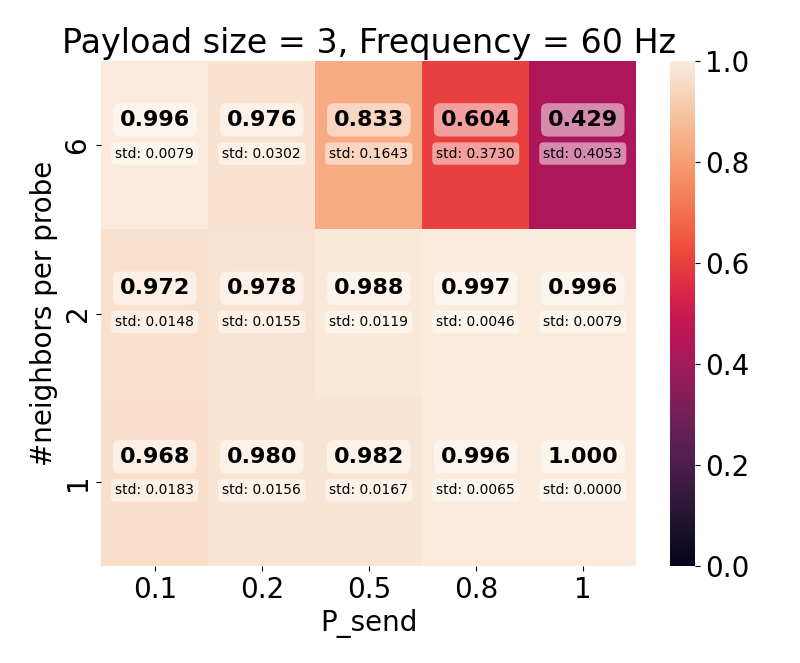}
    \includegraphics[width=0.33\linewidth]{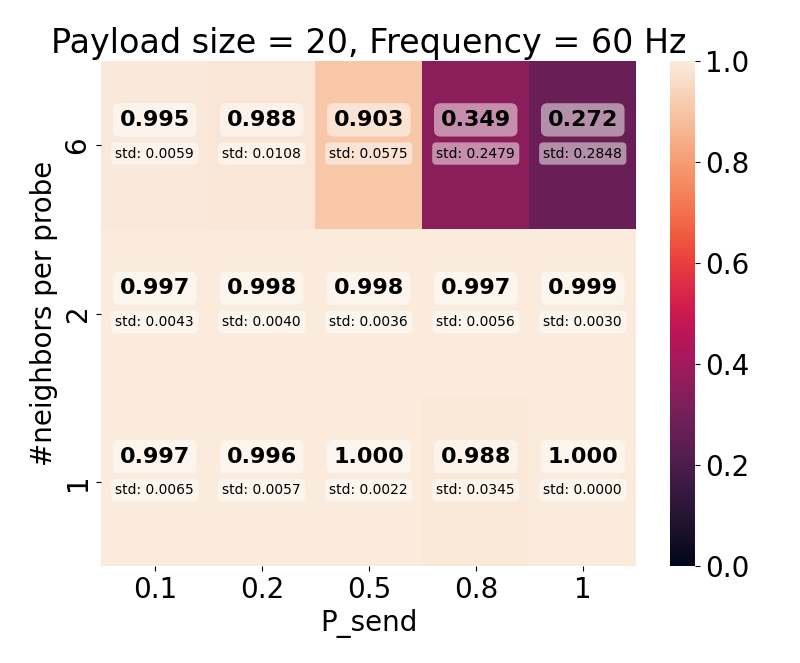}
    \includegraphics[width=0.33\linewidth]{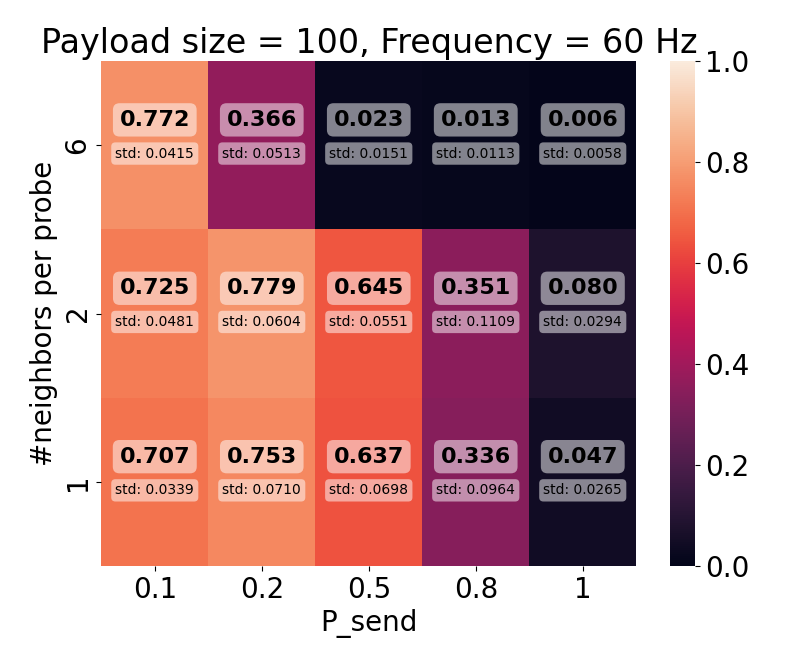}
    \caption{Probability of successful message transmission for different parameters - Varying $p\_send$ the probability of sending, $m$ the number of neighbors around the probe, and the message payload size (3, 20 or 100 bytes). Experiments on two probes sending 100 messages and determining how many were received from the other. Expected time to send 100 messages: from 100 steps ($p\_send = 1$, 1 message per step) to 1000 steps ($p\_send = 0.1$, 1 message per 10 steps)}.
    \label{fig:successful_transmission_proba}
\end{figure*}

\begin{figure*}
    \centering
    \includegraphics[width=1.0\linewidth]{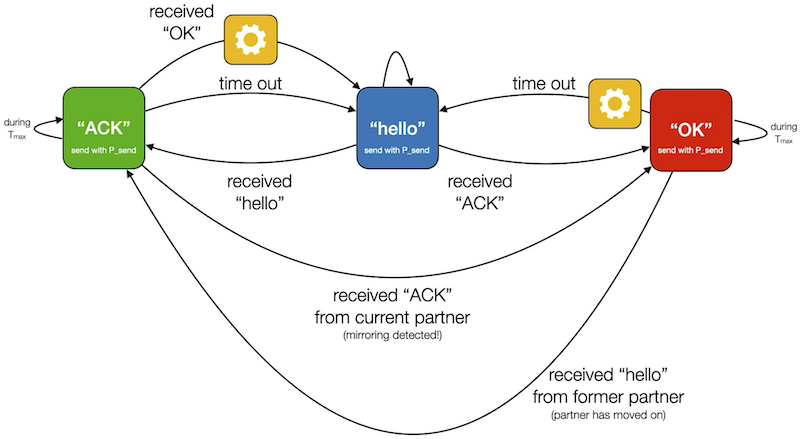} 
    \caption{Handshake Finite State Machine using the Banshee protocol.}
    \label{fig:handshake_FSM}
\end{figure*}

\begin{figure}
    \centering
    \includegraphics[width=0.9\linewidth]{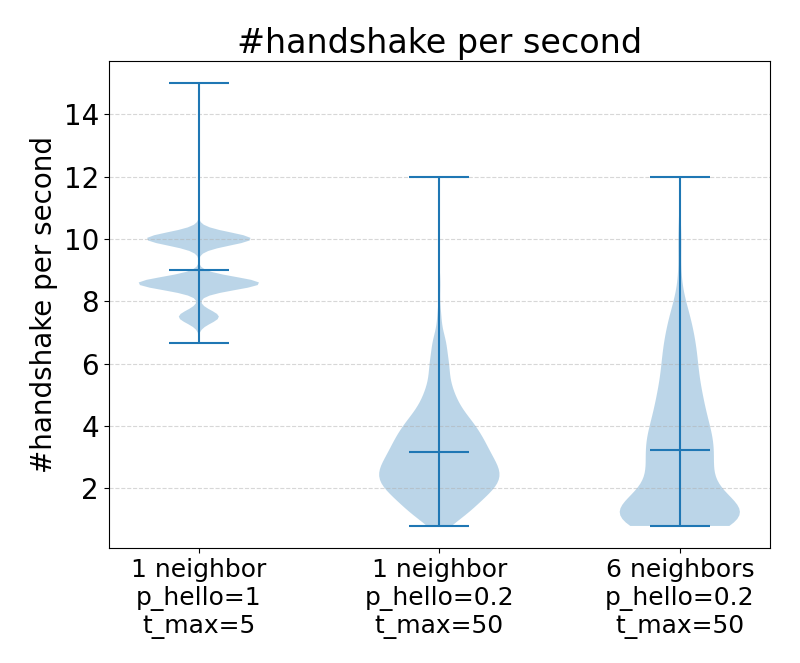}
    \caption{Distribution of handshake rates under various settings. Left: optimal values for sparse density. Middle: optimal values for maximum packing, tested in sparse density. Right: optimal setting for (and tested with) maximum packing. 
    $100$ handshakes with $2$ robots (i.e., $200$ data points) in varying conditions.
    }
    \label{fig:handshakepersec}
\end{figure}

Infrared (IR) communication is the primary medium for interaction among Pogobots. In the ideal pairwise case (one sender, one listener), the communication channel is fast and reliable. As shown in Fig.~\ref{fig:IRmaps}-left, messages are transmitted at 60Hz with a payload of 64~bytes, achieving nearly perfect delivery up to 10cm. Higher frequencies can be reached by reducing the payload size (up to 100–120Hz for a payload under 20 bytes), while using the maximum payload size of 391~bytes yields stable communication at 30~Hz.

The IR system uses four sensors distributed around the robot body, with overlapping fields of view (Fig.~\ref{fig:IRmaps}-right). This configuration creates eight angular sectors, allowing each robot to infer the relative position of a sender, and to coarsely estimate its orientation. This directional information can be used for behaviors requiring alignment or orientation-dependent positioning.

When communication occurs in denser settings where robots act as both senders and listeners, packet collisions become possible. Fig.~\ref{fig:successful_transmission_proba} illustrates how performance of a focal robot degrades with increasing neighborhood size, message payload and larger probability of sending a message ($p_{send}$), as collisions become more frequent. The practical upper bound in our setup is six neighbors (hexagonal packing).
It is important to note that even with robots using $p_{send}=1$ (i.e., bust-mode), packet loss is lower than what could be expected, especially when the payload is smaller. Though robots run at the same frequency (here: 60Hz), they may not send messages at the same time, i.e., the robots' discrete time slots are not synchronized. Moreover, messages are faster to transmit ($<<1/60s$) than the time between control updates, which means only part of the time lapse between two steps is used for transmission. As a consequence, shorter messages implies a lower the probability of packet collision. Conversely, longer messages and/or denser neighborhood increase the probability of packet collision.

We introduce the Banshee protocol (details in \anonym{Annex~\ref{annex:com}}) to mitigate packet loss in the worst case, providing that neither collision detection (conflict cannot be detected) nor collision resolution (acknowledgement is not considered) are realistic options when highly motile robots using a random-access channel are considered. Banshee is a simplistic probabilistic protocol that regulates message transmission by setting the probability of sending at each control cycle. 
Since robots move continuously and cannot rely on persistent links, the protocol adopts an epidemic style: messages are sent without certainty of reception, but with high enough likelihood to ensure effective propagation across the swarm.

The Banshee protocol uses a probability $p_{\text{send}}$ to transmit a message at each control step. The optimal value is $1/N$, where $N$ is the number of interacting neighbors (incl. the focal robot), balancing transmission attempts and silence periods. In practice, $p_{\text{send}}$ is chosen based on swarm density: in dilute conditions two robots interacting give $p_{\text{send}}=0.5$, while in dense hexagonal packing with six neighbors the theoretical optimum is $\approx0.14$. Experiments showed that real Pogobots, thanks to their four directional IR receivers (with dedicated electronics) and lack of global synchronization, can tolerate slightly higher values, so we set $p_{\text{send}}=0.2$ to ensure communication in both dilute and dense environment.

For cases where guaranteed exchange is required (i.e., two robots should have a very strong confidence on the symmetry of an exchange), we implement a handshake protocol (Fig.~\ref{fig:handshake_FSM}). Built on top of Banshee, it employs a finite-state automaton to synchronize sender and receiver. While full certainty cannot be achieved due to the impossibility of perfect consensus (Byzantine generals paradox~\cite{lamport2019byzantine}), the protocol establishes a strong confidence that two robots have successfully exchanged useful payloads. This mechanism is particularly relevant for tasks such as diffusion, where conservation laws (e.g., of virtual matter) must be respected.

Experimental results are summarized in Fig.~\ref{fig:handshakepersec}. The handshake process requires multiple exchanges between robots, but remains robust across densities. The first column shows the optimal configuration for low-density (“gaseous”) conditions, the third column corresponds to high-density configurations (hexagonal packing with six neighbors), and the middle column gives the cost of applying high-density parameters in a low-density regime.
While $p_\text{hello}$ is the same as Banshee's $p_{send}$, parameter for timeout was first estimated from the idealized model, then refined through systematic experimental calibration (details in \anonym{Annex~\ref{annex:com}}). 


\section{Software and Hardware Support Tools}
\label{sec:ext}

\subsection{PogoTracker: Visual tracking Systems}
\label{sec:pogotracker}

To facilitate the analysis of swarm robotics experiments using Pogobots, a video processing tracking pipeline is available at~\anonym{\url{https://github.com/keivan-amini/pogotrack}}. This package includes a set of 3D-printable parts that enable the extraction of Pogobots' dynamical variables — such as $(x, y, \theta)$ on a plane, where $\theta$ denotes the robot’s orientation — from recorded video experiments (see Fig.~\ref{fig:tracking}-a). In addition, the pipeline provides tools to process, clean, extract and plot physical variables derived from these recordings, using a set of tunable video-processing parameters and a command-line interface. 


\subsection{Pogosim: a fast API-compatible Simulator}

\begin{figure}[htbp] 
    \centering
    \includegraphics[width=0.9\linewidth]{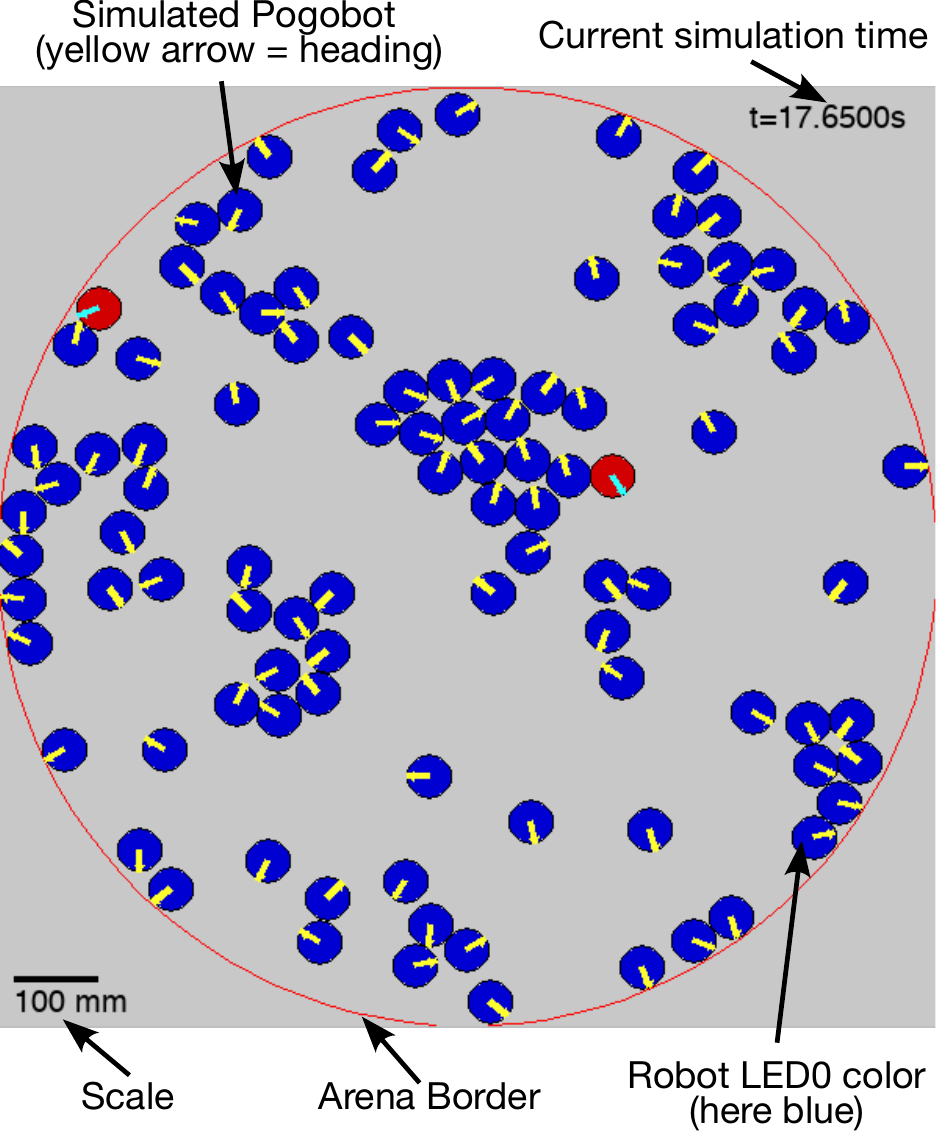} 
    \caption{Snapshot of the Pogosim simulator.}
    \label{fig:pogosim}
\end{figure}

Pogosim~\anocite{\cite{cazenille2025pogosimsimulatorpogobot}} is a simulator that mirrors the robot hardware and APIs, so that the same C user code runs unmodified in simulation and with real robots (see Fig.~\ref{fig:pogosim} for a snapshot). 
The simulator provides a realistic modeling of the communication protocol and locomotion dynamics (using data presented in earlier Sections), in order to minimize the reality gap when code is transferred to the real robots. Pogosim is available at~\anonym{\url{https://github.com/Adacoma/pogosim}}.
It is written in C/C++, using Box2D for physics and integrates pseudo-realistic IR communication system (incl. packet loss), batch execution tools, and parameter optimization routines. It is inspired from the Kilombo simulator~\cite{jansson2015kilombo}. In terms of performance, a high-end laptop (Core i7-11900H@2.3GHz) takes from 2.76~sec (no GUI, no data log) to 4.2~sec (GUI, extensive log) to simulate $100$ Pogobots with a controller frequency of 60Hz performing run\&tumble in a circular arena of $1.5m^2$ during 1 minute, i.e. a 14x to 21x acceleration factor (average over $5$ runs for each setup).

\subsection{Hardware extensions}
\label{sec:hwext}

\begin{figure}[htbp]
    \centering
    \includegraphics[height=2.725cm]{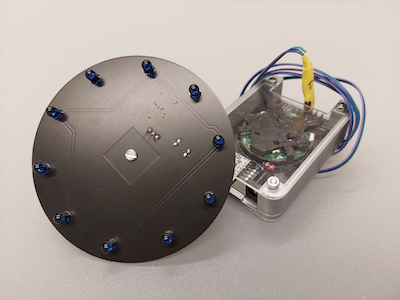}%
    \hspace{0.4cm}%
    \includegraphics[height=2.725cm]{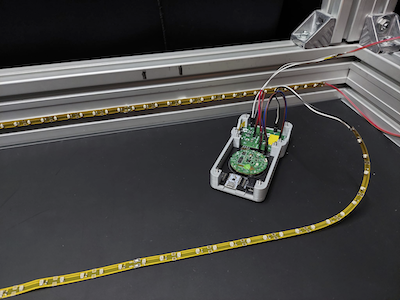}%
    \hfill
    \includegraphics[height=3.5cm]{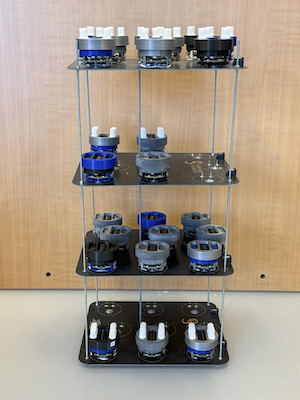}%
    \hspace{0.4cm}%
    \includegraphics[height=3.5cm]{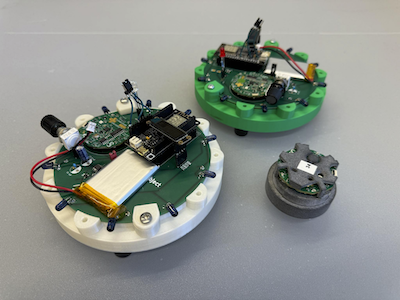}
    \caption{(a) Pogoshower; (b) Pogowall, (c) 4 stacked Pogochargers (with charging Pogobots) ; (d) A Pogobot, a movable Pogobject and a static Pogobject.}
    \label{fig:extensions}
\end{figure}

Hardware extensions are shown in Figure~\ref{fig:extensions} and described hereafter.

\textbf{Pogoremote:} a dedicated board used to send signals to Pogobots through Infrared communication. The Pogoremote is connected through a cable to a nearby computer, so that the experimenter can send commands or programs to Pogobots. The Pogoremote also comes in the form of a \textbf{Pogoshower}, a 100mm diameter device with 10 IR emitters that can be manipulated by hand to program Pogobots simultaneously within a directed cone within a range of approx.~50~cm. It can also be used to send user-defined signals to the Pogobots in range (e.g., to start or stop an experiment). 

\textbf{Pogowall}: an IR-emitting wall, which consists of a LED strip attached to a physical wall, that can transmit messages to nearby Pogobots. In its simplest form, a Pogowall continuously broadcasts a signal that reveals its presence to nearby robots. A wall identifier can be included in the signal, so that Pogobots can distinguish walls from one another. Pogowalls can also be used to perform mass programming of nearby robots, as each is actually embedding a Pogoremote board.

\textbf{Pogocharger}: An efficient, scalable charging solution. Each Pogocharger is able to charge up to 9 robots in less than~2~hours. Pogochargers can be stacked on one another to minimize space requirement.

\textbf{Pogobject}: Both static and movable objects are equipped with IR communication, allowing dynamic interaction experiments, such as obstacle avoidance, object pushing, and collective transport. 

\section{Conclusion}

The Pogobot platform offers a versatile, cost-effective solution for swarm robotics and active matter research thanks to its enhanced communication capabilities, flexible locomotion options, and extensible architecture.
The open-source and open-hardware approach encourages community contributions and adaptations to specific needs. Firstly, all parts of the exoskeleton (incl. the robot-ground interface) can be easily redesigned using off-the-shelf inexpensive 3D printers available widely. Secondly, the fully open source nature of the project allows for more in-depth modifications for anyone with low-level engineering skills.


\section*{Acknowledgments}

\invisible{This work is supported by \anonym{the MSR and SSR projects funded by the Agence Nationale pour la Recherche under Grant No ANR-18-CE33-0006 and ANR-24-CE33-7791.}. }
Disclosure: English is not our first language, ChatGPT~5 was used for improving part of the main text. \invisible{\anonym{We thank the following people for their help and suggestions thoughout the years: Lyna Saoucha, Sama Satariyan, Mika Ito, Baris Kaftancioglu, Paul Tiberiu Iordache, Salman Houdaibi, Lara Polachini. We also thank early adopters: Nathanael Aubert-Kato (Ochanomizu University, Japan) and Amine Boumaza (Université de Lorraine, France).}}



\bibliographystyle{IEEEtran}
\bibliography{nicolas} 


\clearpage        
\onecolumn        
\appendix

\invisible{
\appendix

\section{Annex: Communication}
\label{annex:com}

Message sent by the Pogobots have a length of maximum 400 bytes. A message can be sent from all $4$ IR emitters, or just a subset. Messages can be of two types: either short or long messages. Short messages have a minimal header that contains the identifier of the sender and a message ID. Long messages extend this header with information on which IR emitter is used to send the message, which makes it possible to collect information on relative orientation between the robot sender and receiver (which also get information about which of its IR receiver(s) collected the message). 

Message are structure as follow: start sequence (1 byte), Header (3 to 8 bytes), CRC (4 bytes), end sequence (1 bytes). The payload is 386 bytes for long messages, or 391 bytes for short messages. In practical, the payload contains just what is needed (i.e., from 1 single byte to maximum payload).

The Banshee communication protocol is implemented so as to maximize the number of messages sent \textit{and} received, which implies a trade-off between maximizing the number of messages sent and minimizing the risk of packet loss due to collisions. The constant dynamic reconfiguration of the robot swarm, as well as the lack of \textit{reliable} collision detection, result in proposing a solution where the probability of sending a message is the sole mean to achieve efficient communication.

We define $p_{send}$, the probability that a robot sends a message at one timestep, which is valid for all robots in the swarm. There is a tradeoff between speed (higher $p_{send}$ is desired) and packet loss due to collision (lower $p_{send}$ is preferred). Then:

\noindent
For agent $i$, with a probability to send a message at one timestep $P_{send}$ let
\[
p_{solo} \;=\; P_{\text{send}} \,(1 - P_{\text{send}})^{N-1}
\]
be the probability that, in a given controller step, agent $i$ sends a message while all the other $N-1$ agents remain silent.

\begin{enumerate}
  \item Average waiting time until success (ie. robot A was alone to send a message):
  \[
  T^{i}_{\text{avg}} \;=\; \frac{1}{p_{solo}}.
  \]

  \item Guaranteed success within tolerance $P_{\text{tol}}$:
  \[
  \Pr\!\left(T^i \leq T \right) = 1 - (1 - p_{solo})^T,
  \]
  hence the maximum number of steps to succeed with probability at least $P_{\text{tol}}$ is
  \[
  T^{i}_{\max}(P_{\text{tol}}) \;=\; \frac{\ln(1 - P_{\text{tol}})}{\ln(1 - p_{solo})}.
  \]
\end{enumerate}

\medskip
\noindent
The optimal sending probability, which maximizes $p_{solo}$, is then:
\[
P_{\text{hello}}^{\star} = \frac{1}{N},
\quad\;\;
p_{solo}^{\star} = \frac{1}{N}\left(1 - \frac{1}{N}\right)^{N-1},
\quad\;\;
T^{i}_{\text{avg}} = \frac{1}{p_{solo}^{\star}}.
\]

In terms on implementation, we have to consider the optimal value of $P_send$ in terms of how the number of robots in interaction. Such number depends on how dense the swarm is, from a dilute gas-like system where only two robots interact at one time, to a dense packing resembling a crystal where each robot has 6 neighbors (hexagonal packing). 

\begin{itemize}
\item dilute phase, with 1 focal robot interacting with 1 other robot at a specific time): $P_{send}=0.5$, with expected $T_{avg}=4$ steps, and $T_{max}=16$ steps for 99\% tolerance ($T_{max}=24$ for 99.9\%);
\item dense phase, with 1 focal robot surrounded by 6 robots: $P_{send}=0.1429$, with expected $T_{avg}=17.6$ steps, and $T_{max}=39.6$ steps w/ 99\% tolerance ($T_{max}=79$ for 99.9\%).
\end{itemize}

In practical, we prioritize maintaining communication even in the worst case (the dense phase), and should start with a value of $P_{send}=0.1429$. 

However, extra tuning performed on the robots in real condition allowed to increase this value up to $P_{send}=0.2$. This is due to the fact that actual packet loss is to be considered for each of the four Pogobot's IR receivers. Each covers approx. $90$ deg. and receives in the worst case messages from $3$ to $4$ neighboring robots. In that case 
dense phase, with 1 focal robot surrounded by 6 robots, we consider that each of the focal Pogobot's IR receiver faces a maximum of $4$ robots, then: $P_{send}=0.2$, with expected $T_{avg}=12.2$ steps, and $T_{max}=54$ w/ 99\% tolerance ($T_{max}=81$ for 99.9\%). With a controller frequency of $60Hz$, Pogobots are then expected to send approx.~$12$ \textit{useful} messages per second, with a payload of up to $391$ bytes.

\section{Annex: Locomotion}
\label{app:technical}

The studied physical observables have been extracted by using the following methods. \\ 
In noisy and stochastic environments, the linear velocity of a particle can be estimated from the mean square displacement (MSD):  

\[
\text{MSD}(\tau) = \big\langle \, \lVert \mathbf{r}(t+\tau) - \mathbf{r}(t) \rVert^2 \, \big\rangle_t ,
\]

where $\tau$ is the lag time and $\mathbf{r}(t) = (x(t),y(t))$ the particle position.  
At short times, trajectories are approximately ballistic, so the MSD scales quadratically:  

\[
\text{MSD}(\tau) \;\approx\; v^2 \tau^2 .
\]

Thus, a linear regression of $\text{MSD}(\tau)$ versus $\tau^2$ over the early-time regime yields the squared velocity, from which we can extract the root. Only a fraction of the smallest lag times is used in the fit to ensure the ballistic regime is captured.
\\
The exact same MSD approach on variable $\theta$ has been used to estimate the angular velocity $\Omega_{MSD}$.

The radius of curvature $R$ was computed by fitting circles to individual trajectories using least-squares optimization. The fitted radius provided a quantitative measure of how straight the robots moved under each locomotion condition.  
\section{Annex: Hardware}
\label{annex:hw}
This document provides a technical description of the Pogobot mobile robot.

The Pogobot project is an open source / open hardware initiative the Institut des Systèmes Intelligents et de Robotique (Sorbonne Université, CNRS), with funding from the Agence Nationale pour la Recherche under Grant No ANR-18-CE33-0006. This project aims to produce small robots for smarm and collective robotics. 

The robot is composed of 2 PCBs ("brain" and "belly") connected together inside a PLA capsule. The "brain" of the robot is a FPGA that implements a System on Chip (LiteX). This SoC has a RiscV softcore CPU and peripherals connected to it.


\subsection{Functionalities}
The robot features various sensors and actuators (see also Section~\ref{sec:design} and Fig.~\ref{fig:pogobot_general}):
\begin{itemize}
    \item 4x Infrared I/O (per direction: 2 coupled emitters, one receiver);
    \item 3x photosensors (2 at the front, 90 deg. appart; 1 at the back);
    \item 1x 6-axis Inertial Measurement Unit
    \item 5x RGB LEDs (around the robot)
    \item 3x vibrating motors (2 on the sides and 1 in the back)
    \item 1x 3.3V LiFePO4 CR2 rechargeable battery
\end{itemize}


\subsection{Overview}

The different parts of the robots and the view of the fully assembled robots are shown in Figure~\ref{fig:robotall}.

A Robot is composed of 2 PCBs, 3 motors and several 3D-printed parts:
\begin{itemize}
\item The main PCB is named the Head at the top. It contains the FPGA that is the "brain" of the robot, all the sensors and 1 LED.
\item The second PCB is named the Belly and is connected below the head. It manages the power management (battery and power chip), the connection with the motors and contains 4 LEDs.
\item The different 3D printed parts are named capsule (for the one against the Belly PCB) and the skirt that locks the capsule. In this example, the skirt holds the toothbrush teeth that allows the robot to move. Other skirts have been designed and tested, such as different positioning for the toothbrush, flexible legs, metallic legs, and even 2-wheel skirt.
\end{itemize}

The robot moves using its vibrating motors. In the example shown here, vibrations make the robot performs micro jumps, and the whole system goes straight forward or turn, depending on the particular vibrating speed of the different motors.

\begin{figure*}[t]
\centering

\begin{subfigure}{0.32\textwidth}
  \centering
  \includegraphics[width=\linewidth]{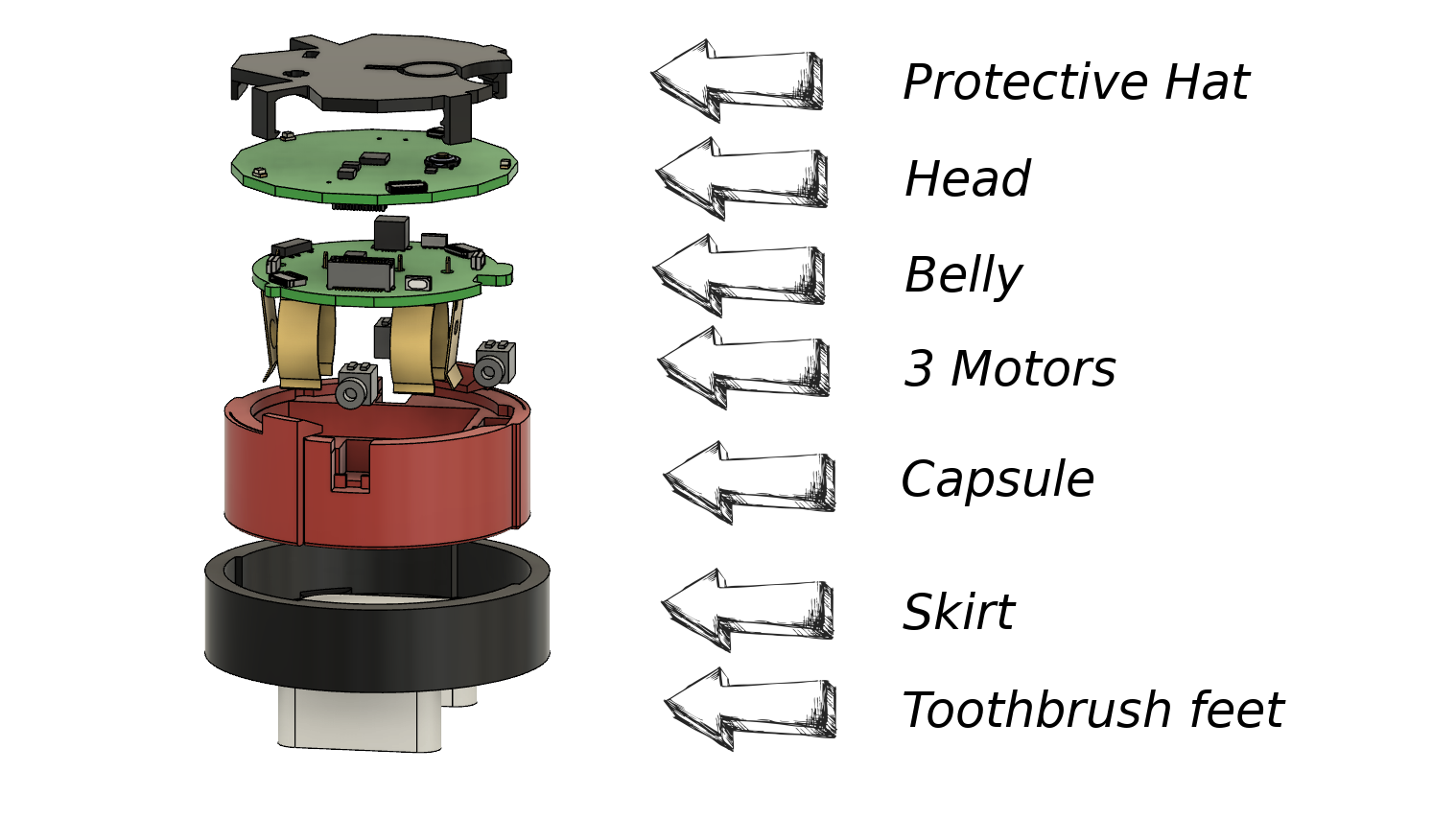}
  \caption{Pogobot exploded view}
  \label{fig:eclate}
\end{subfigure}
\hfill
\begin{subfigure}{0.32\textwidth}
  \centering
  \includegraphics[width=\linewidth]{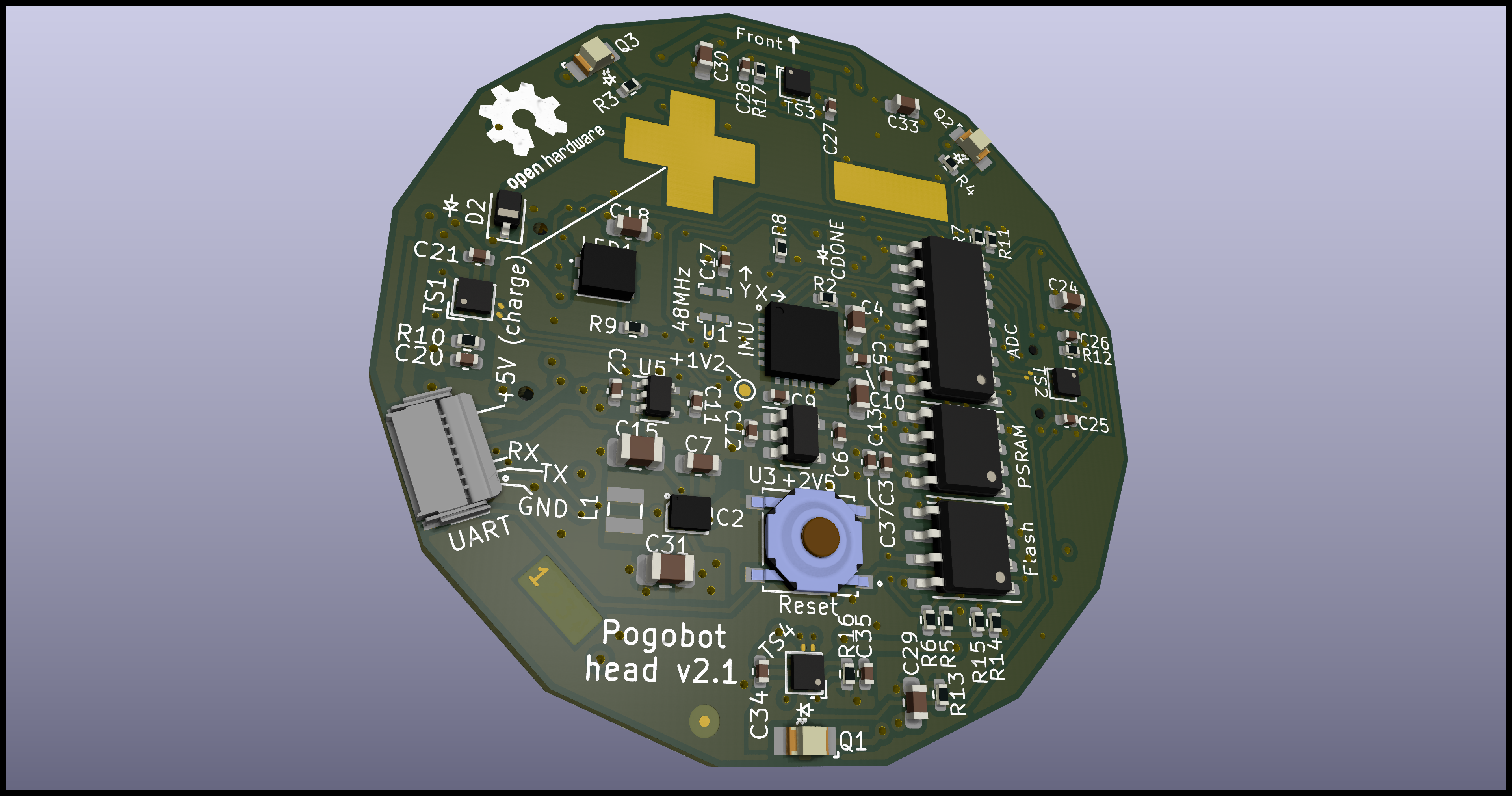}
  \caption{Pogobot top head}
  \label{fig:head_top}
\end{subfigure}
\hfill
\begin{subfigure}{0.32\textwidth}
  \centering
  \includegraphics[width=\linewidth]{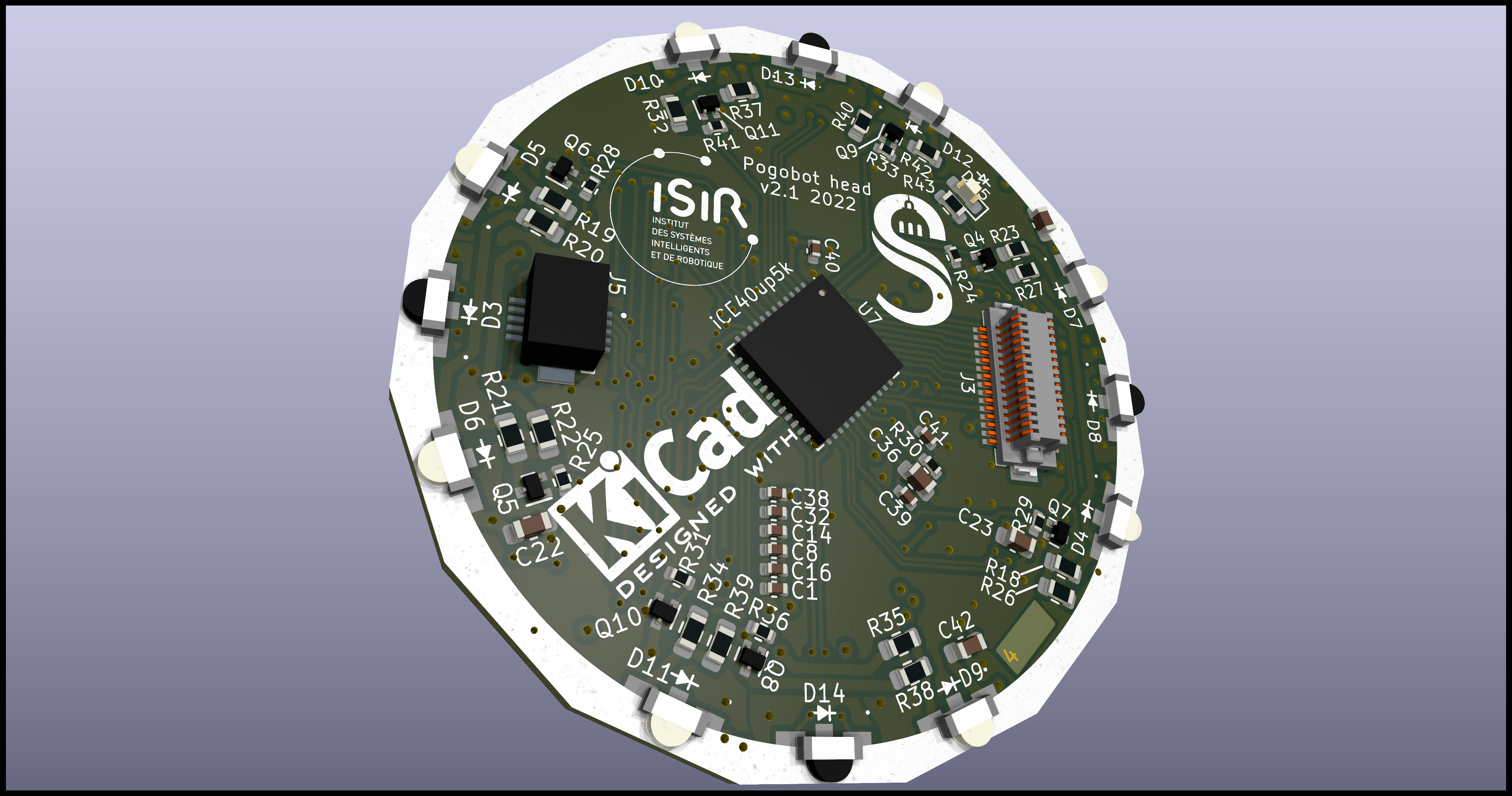}
  \caption{Pogobot bottom head}
  \label{fig:head_bottom}
\end{subfigure}

\medskip

\begin{subfigure}{0.32\textwidth}
  \centering
  \includegraphics[width=\linewidth]{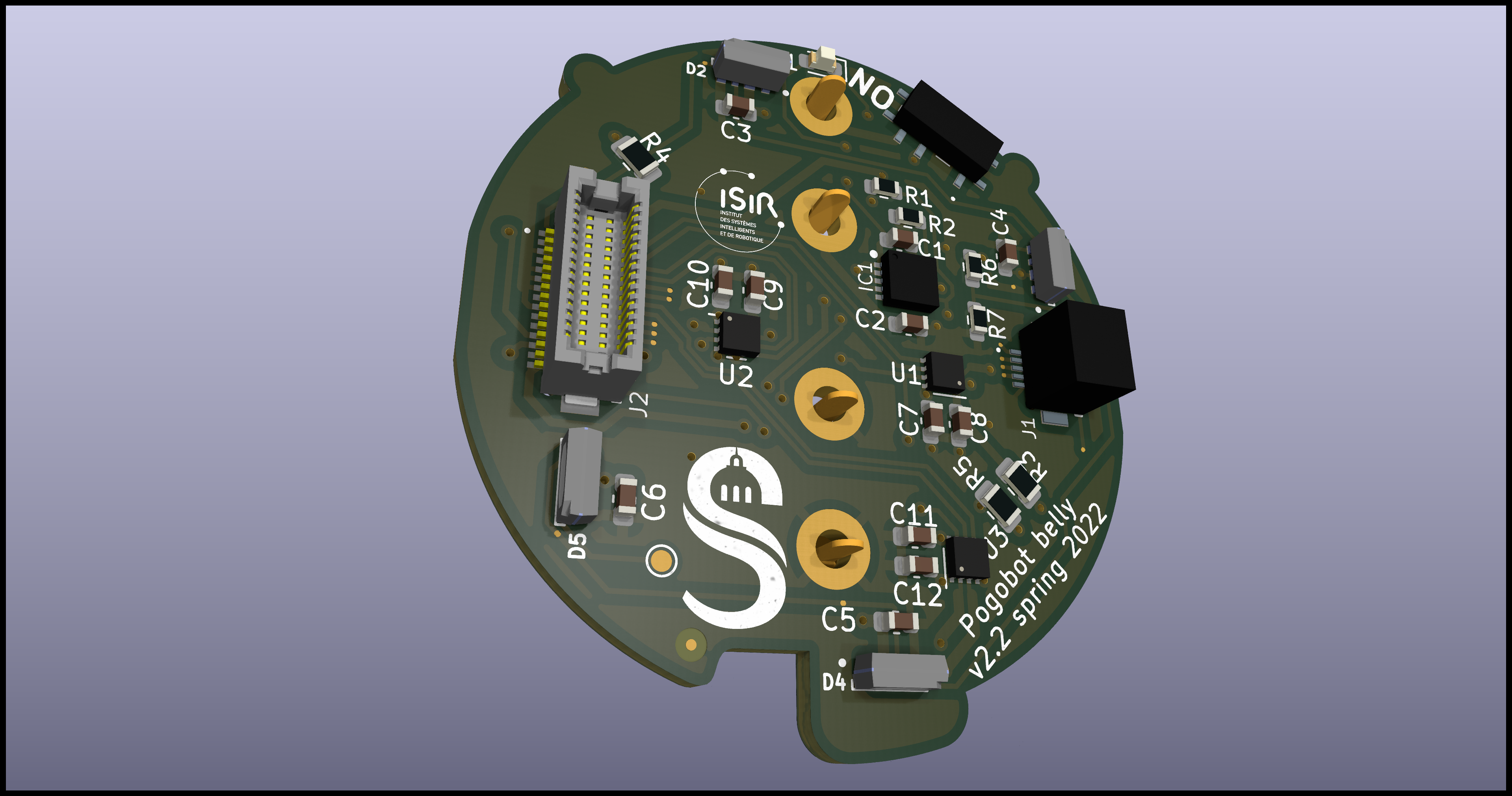}
  \caption{Pogobot top belly}
  \label{fig:belly_top}
\end{subfigure}
\hfill
\begin{subfigure}{0.32\textwidth}
  \centering
  \includegraphics[width=\linewidth]{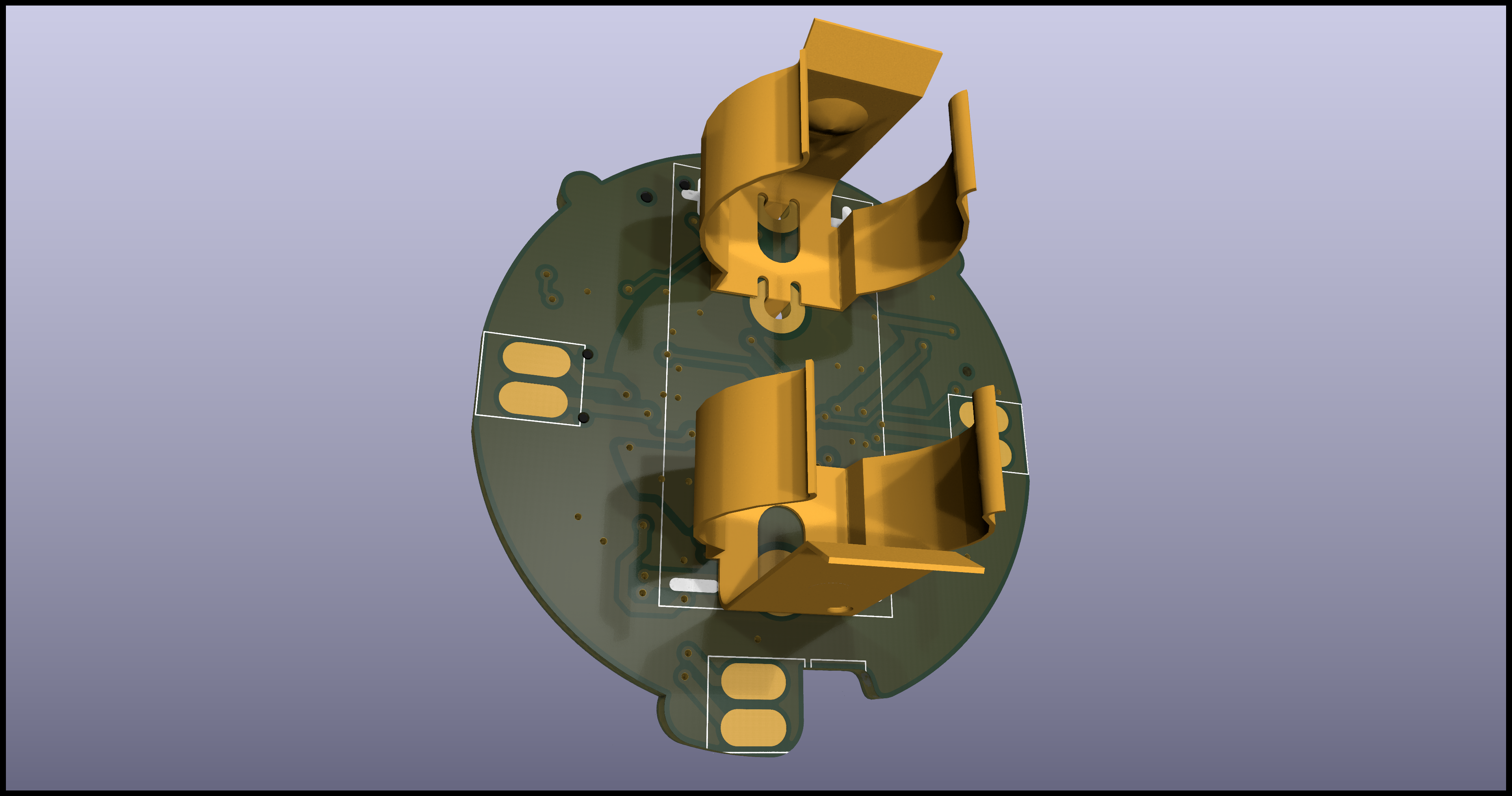}
  \caption{Pogobot bottom belly}
  \label{fig:belly_bottom}
\end{subfigure}
\hfill
\begin{subfigure}{0.32\textwidth}
  \centering
  \includegraphics[width=\linewidth]{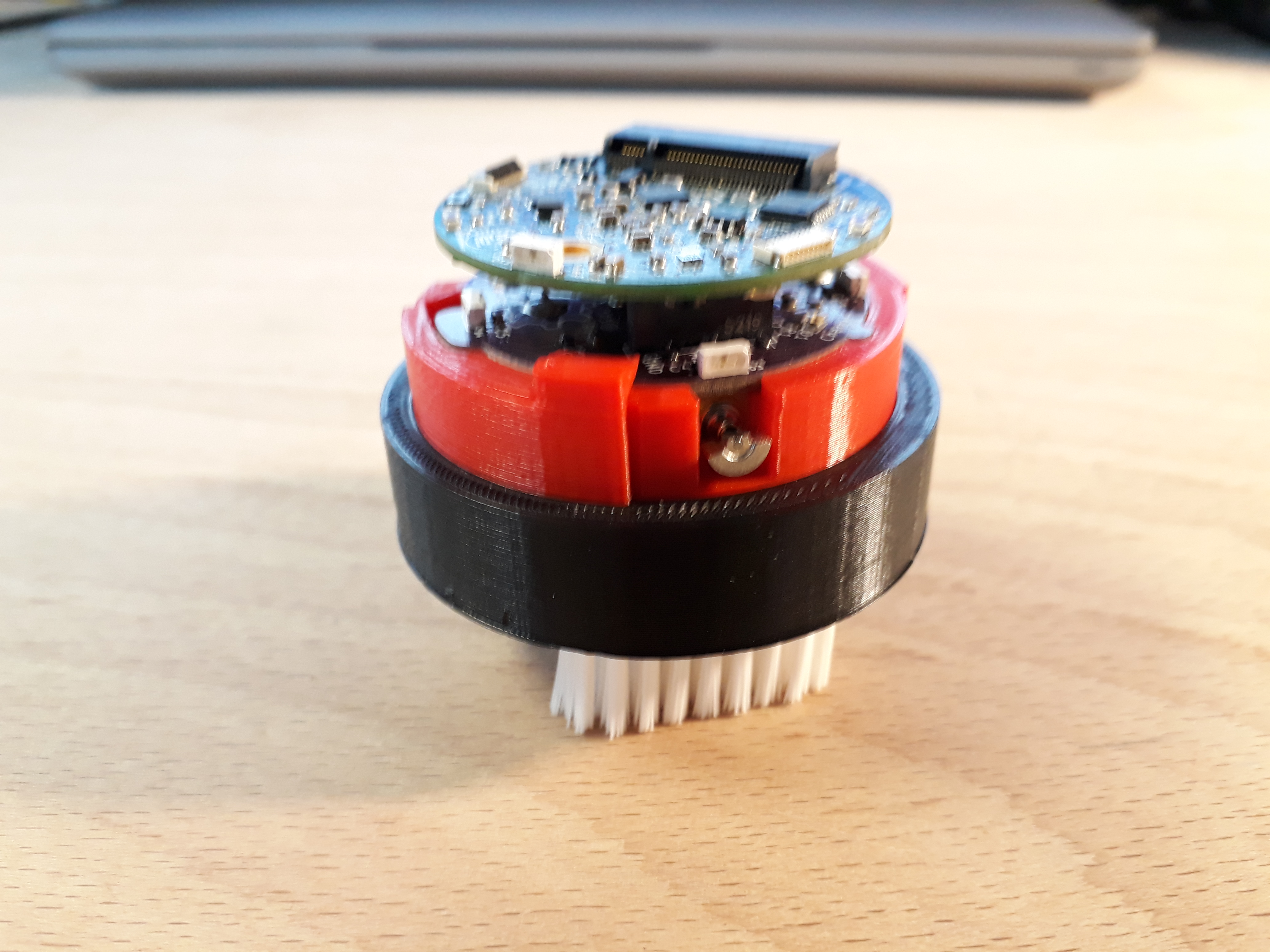}
  \caption{Fully assembled robot}
  \label{fig:robotall}
\end{subfigure}

\caption{Robot mechanics showing exploded view, head and belly (top/bottom), and fully assembled configuration.}
\label{fig:robot_view}
\end{figure*}


\subsection{Pogobot States}

Figure~\ref{fig:state_machine} shows the possible states of a Pobobot, and the color code displayed to the human user using the overhead LED.

\begin{figure}
\centering
\includegraphics[width=1.0\linewidth]{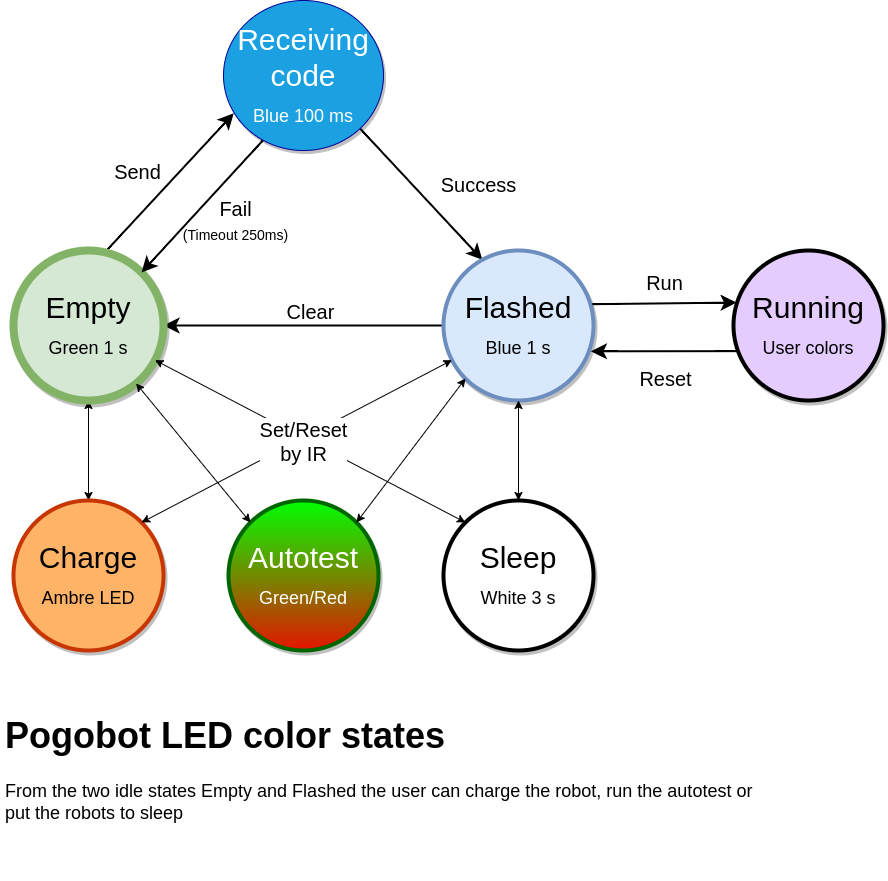}
\caption{State diagram of Pogobot LED color codes.}
\label{fig:state_machine}
\end{figure}


\subsection{Details}
The project can be divided into 2 parts. 
\begin{itemize}
    \item The Hardware and the Gateware: the PCBs, the mechanical parts and the code produced for the FPGA.
    \item The Software: the code produced for the softcore CPU inside the FPGA 
\end{itemize}


\subsubsection{Hardware / Gateware}

The figure \ref{fig:hardall} shows the overview schematic of the robot.

\begin{figure*}
\centering
\includegraphics[width=1\linewidth]{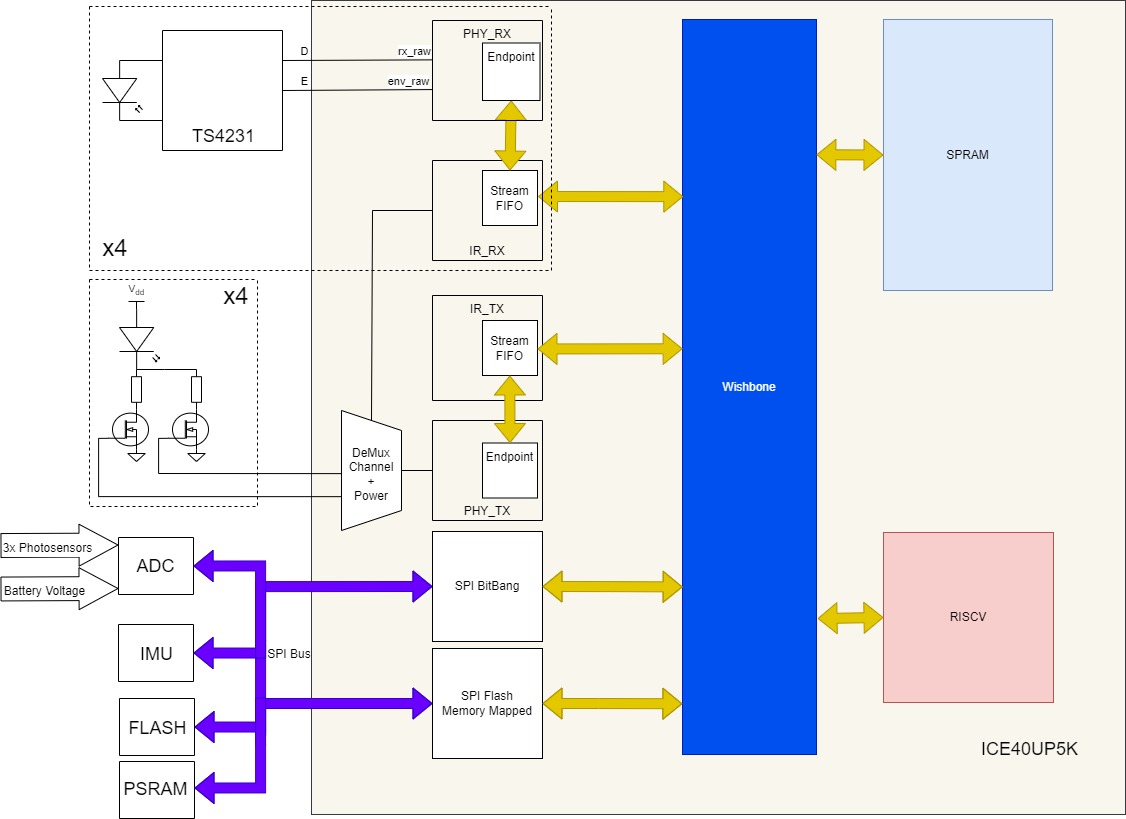}
\caption{Hardware overview schematic}
\label{fig:hardall}
\end{figure*}

The robot is implemented around a FPGA ICE40, a TS4321 (IR reception manager) and different sensors connected through a SPI Bus. 

Inside the FPGA, a RiscV softcore is available and everything is connected to a Whishbone Bus that insures the communication. The SRAM is the one of the FPGA. Some hardware are also implemented to interface the inputs/outputs with the Whishbone interconnection.

All the gateware is generated using the \href{https://github.com/enjoy-digital/litex}{LiteX}. This framework allows the user to generate all the code (gateware) for the FPGA from Python and in the same time the low level software interface.

\section{Annex: Software}
\label{annex:sw}

We develop an "easy to use" C library in order to abstract the robot constraints and allows to access all the sensors and actuators with simplicity. 
A SDK is available to be able to concentrate on the software code but as a opensource project everything is available to change. 
A bunch of examples are available to test the possibilities.

\subsection{Examples}

\paragraph{Led Control}

Figure~\ref{lst:ledcontrol} shows how to change LED no.0 to a random color every 100ms.

\begin{figure*}[t]
\centering
\begin{minipage}{0.95\textwidth}
\begin{lstlisting}[language=C,basicstyle=\ttfamily\scriptsize,caption=Led Control example]
#include "pogobot.h"

int main (void)
{
    pogobot_init();
    /* initialize random */
    srand( pogobot_helper_getRandSeed() );

    while(1) 
    {
        pogobot_led_setColor( rand()%255, rand()%255, rand()%255);
        msleep( 100 );
    }
}
\end{lstlisting}
\end{minipage}
\caption{Example for setting LED colors.}
\label{lst:ledcontrol}
\end{figure*}

\paragraph{Motor Control}
The following code starts the right ad left motor at half power.
\begin{lstlisting}[language=C,basicstyle=\ttfamily\scriptsize,caption=Motor Control example]
#include "pogobot.h"

int main (void)
{
    pogobot_init();
    
    pogobot_motor_power_set(motorL, motorHalf);
    pogobot_motor_power_set(motorR, motorHalf);

}
\end{lstlisting}

\paragraph{Infrared Communication}

Figure~\ref{lst:ir_comm} provides an example for sending and receiving messages through IR communication. 

\begin{figure*}[t]
\centering
\begin{minipage}{0.95\textwidth}
\begin{lstlisting}[language=C,basicstyle=\ttfamily\scriptsize]
#include "pogobot.h"

typedef struct my_data_t
{
    uint16_t genome;
    uint16_t fitness;
} my_data_t;

my_data_t my_data = { .genome = 0, .fitness = 0 };

int main(void) {

    pogobot_init();
    // select the IR power
    pogobot_infrared_set_power(pogobot_infrared_emitter_power_twoThird);

    while (1)
    {
        pogobot_infrared_update();
        
        int destinationId = 0x1234;
        my_data.genome = Ox0F;
        my_data.fitness = 4;
        // send Message
        pogobot_infrared_sendMessageAllDirection( destinationId, 
                                                  (uint8_t *)( &my_data ),
                                                  sizeof( my_data ) );

        // take the first received message if it exists
        if ( pogobot_infrared_message_available() )
        {
            message_t mr;
            pogobot_infrared_recover_next_message( &mr );
            my_data_t *recept = (my_data_t *)(&( mr.payload ));
            
            printf("%d; %d \n", recept->genome, recept->fitness);
            
        }
        // clear queue
        pogobot_infrared_clear_message_queue();

        msleep( 100 ); 
    }
}

\end{lstlisting}
\end{minipage}
\caption{Example of sending and receiving messages with Pogobot IR communication.}
\label{lst:ir_comm}
\end{figure*}

\subsection{Softcore Benchmark}

We simulate inference on a Feed-forward Artificial Neural Network with an hyperbolic tangent as an activation function (fig \ref{fig:NNeval}) to provide an idea of performance. The ANN implements real values weights, the number of weights sets the number dimensions. A controller frequency of 60Hz makes it possible to run an ANN with 200-300 weights in real-time.

\begin{figure*}
\centering
\includegraphics[width=0.7\textwidth]{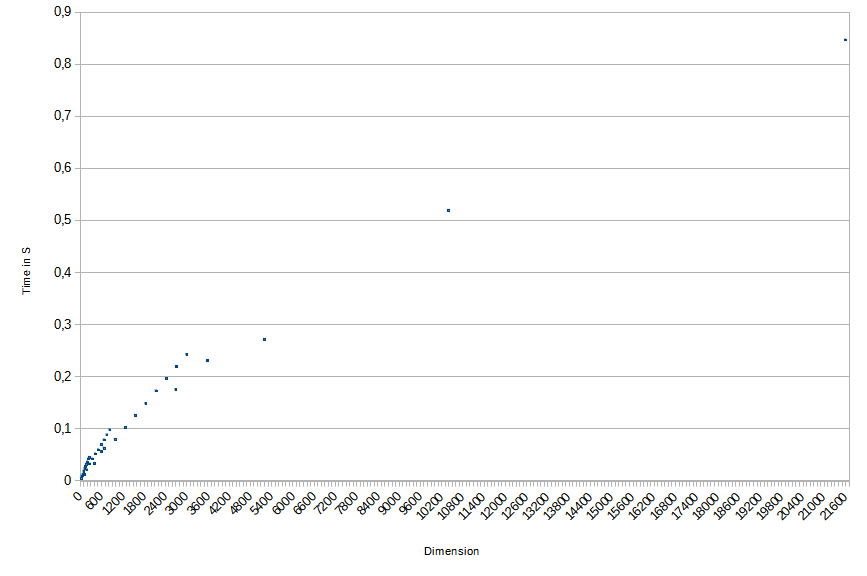}
\caption{On-board Performance of an Artificial Neural Network (\textit{tanh} activation function, feed-forward, real-values)}
\label{fig:NNeval}
\end{figure*}


\subsection{Automatic Calibration}
\label{sec:autocalibration}

An automatic motor calibration procedure was implemented to compensate for asymmetries in wheel alignment and motor response. The algorithm is based on an Extended Kalman Filter (EKF) that estimates the robot’s yaw-rate from inertial measurements and iteratively corrects the motor command asymmetry so that the mean yaw-rate converges toward zero. 

The calibration process consists of repeated short trials. During each trial, the robot is commanded to move forward at a fixed input power while accelerometer and gyroscope data are collected from the onboard IMU. These six-dimensional observations (three axes of acceleration and three of angular velocity) are processed in real time by the EKF, which combines prediction and correction steps. In compact form, the update can be written as

\[
\hat{x}_k = A \hat{x}_{k-1} + v_k, 
\qquad 
\hat{x}_k \;\leftarrow\; \hat{x}_k + K_k \big(z_k - H \hat{x}_k\big),
\]

where $\hat{x}_k$ is the estimated state (including yaw-rate), $z_k$ the IMU observation, $K_k$ the Kalman gain, and $v_k$ the process noise. The filtered estimate of the yaw-rate ($\omega_z$) is used as feedback: if the robot veers left or right, the corresponding motor powers are incrementally corrected. This procedure is repeated for a user-defined number of trials until convergence. The final calibrated motor values can be stored in memory for subsequent use.

The implementation developed for the Pogobot platform is publicly available at: \anonym{
\url{https://github.com/PaulTiberiu/motor_calibration_pogobot/tree/main/examples/pogoKalman}.
}

}

\end{document}